\documentclass[11pt,a4paper]{article}
\usepackage[hyperref]{emnlp2020}
\usepackage{times}
\usepackage{latexsym}
\usepackage{graphicx}
\usepackage{subcaption}
\usepackage{url}
\usepackage{latexsym}
\usepackage{graphicx}
\usepackage{booktabs}
\usepackage{multirow}
\usepackage{mathrsfs}
\usepackage{amsmath}
\usepackage{hyperref}
\usepackage{bbm}
\usepackage{amssymb}
\usepackage{wrapfig}
\usepackage{balance}
\usepackage{url}
\usepackage{bm}
\usepackage{float}
\usepackage{amsmath}
\usepackage{amsthm}
\usepackage{amsfonts}
\usepackage{graphicx}
\usepackage{booktabs}
\usepackage{placeins}
\usepackage{multirow}
\usepackage{epstopdf}
\usepackage{color}
\usepackage{subcaption}
\usepackage[T1]{fontenc}
\usepackage{balance}
\usepackage{enumitem}
\usepackage{xcolor}
\usepackage{balance}
\usepackage{relsize}
\usepackage{subcaption}
\usepackage{enumitem}
\usepackage{wrapfig}
\usepackage{graphics}
\usepackage{multirow}
\usepackage{url}
\usepackage{makecell}
\usepackage{array}
\newcommand{\PreserveBackslash}[1]{\let\temp=\\#1\let\\=\temp}
\newcolumntype{C}[1]{>{\PreserveBackslash\centering}p{#1}}
\newcolumntype{R}[1]{>{\PreserveBackslash\raggedleft}p{#1}}
\newcolumntype{L}[1]{>{\PreserveBackslash\raggedright}p{#1}}
\usepackage{microtype}

\definecolor{darkblue}{rgb}{0, 0, 0.5}
\hypersetup{colorlinks=true,citecolor=darkblue, linkcolor=darkblue, urlcolor=darkblue}

\aclfinalcopy 
\def\aclpaperid{485} 

\title{\method{}: Improving Extraction of Task-relevant Utterances through Integration of Discourse Structure and Ontological Knowledge
}

\author{Sopan Khosla \quad Shikhar Vashishth \quad Jill Fain Lehman \quad \textbf{Carolyn Rose}\\
	\vspace{-4 mm} \\
	Language Technologies Institute \\
	Carnegie Mellon University, USA \\
	{\tt \small \{sopank,svashish,jfl,cprose\}@cs.cmu.edu} \\
}
\date{}

\begin{document}
	\newcommand{\refalg}[1]{Algorithm \ref{#1}}
\newcommand{\refeqn}[1]{Equation \ref{#1}}
\newcommand{\reffig}[1]{Figure \ref{#1}}
\newcommand{\reftbl}[1]{Table \ref{#1}}
\newcommand{\refsec}[1]{Section \ref{#1}}

\newcommand{\add}[1]{\textcolor{red}{#1}\typeout{#1}}
\newcommand{\remove}[1]{\sout{#1}\typeout{#1}}

\newcommand{\m}[1]{\mathcal{#1}}
\newcommand{\bmm}[1]{\bm{\mathcal{#1}}}
\newcommand{\real}[1]{\mathbb{R}^{#1}}
\newcommand{\method}{\textsc{MedFilter}}

\newcommand{\bleuone}{BLEU${_1}$\xspace}
\newcommand{\bleu}{BLEU${_4}$\xspace}

\newcommand{\problem}{}
\newcommand{\problemfull}{}

\newtheorem{theorem}{Theorem}[section]
\newtheorem{claim}[theorem]{Claim}

\newcommand{\reminder}[1]{\textcolor{red}{[[ #1 ]]}\typeout{#1}}
\newcommand{\reminderR}[1]{\textcolor{gray}{[[ #1 ]]}\typeout{#1}}

\newcommand{\tensor}{\mathcal{X}}
\newcommand{\Real}{\mathbb{R}}

\newcommand{\tuples}{\mathbb{T}}

\newcommand\norm[1]{\left\lVert#1\right\rVert}

\newcommand{\note}[1]{\textcolor{blue}{#1}}

\newcommand*{\Scale}[2][4]{\scalebox{#1}{$#2$}}%
\newcommand*{\Resize}[2]{\resizebox{#1}{!}{$#2$}}%

\def\mat#1{\mbox{\bf #1}}
\newcommand{\cev}[1]{\reflectbox{\ensuremath{\vec{\reflectbox{\ensuremath{#1}}}}}}

	\maketitle
	\begin{abstract}
		Information extraction from conversational data is particularly challenging because the task-centric nature of conversation allows for effective communication of implicit information by humans, but is challenging for machines. The challenges may differ between utterances depending on the role of the speaker within the conversation, especially when relevant expertise is distributed asymmetrically across roles. Further, the challenges may also increase over the conversation as more shared context is built up through information communicated implicitly earlier in the dialogue.  
		In this paper, we propose the novel modeling approach \method{}, which addresses these insights in order to increase performance at identifying and categorizing task-relevant utterances, and in so doing, positively impacts performance at a downstream information extraction task. We evaluate this approach on a corpus of nearly 7,000 doctor-patient conversations where \method{} is used to identify medically relevant contributions to the discussion (achieving a $10\%$ improvement over SOTA baselines in terms of area under the PR curve). Identifying task-relevant utterances benefits downstream medical processing, achieving improvements of $15\%$, $105\%$, and $23\%$ respectively for the extraction of symptoms, medications, and complaints.
		
	\end{abstract}

	\section{Introduction}
\label{sec:intro2}
\begin{figure*}
	\centering
	\includegraphics[width=\textwidth]{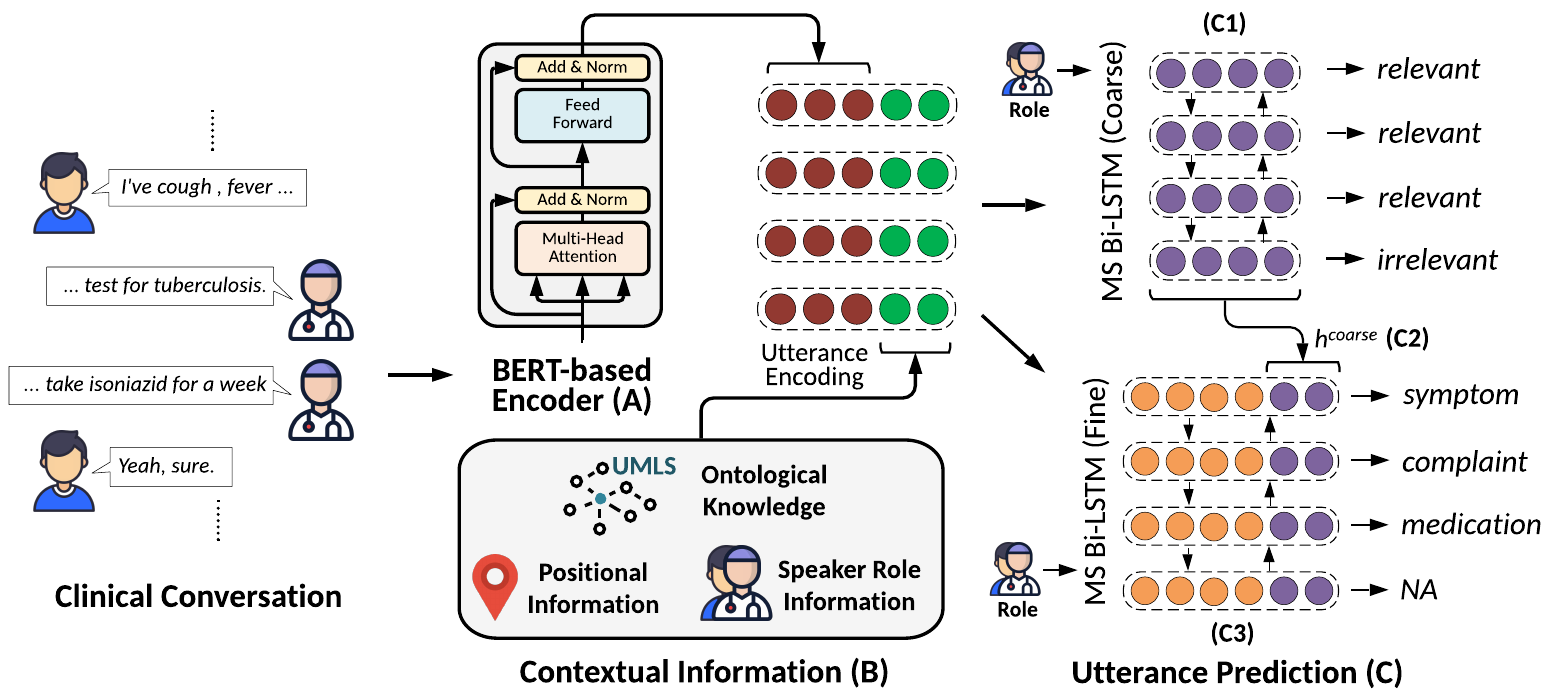}
	\captionsetup{font=small,labelfont=small}
	\caption{\label{fig:workflow} Overview of \method{}. \method{} first encodes each utterance of the given conversation using a BERT-based encoder (A). The obtained utterance embedding is concatenated with contextual information like speaker role, position of utterance in the conversation, and ontological knowledge (B). This is then fed to a MS-BiLSTM (C1) for medical relevance identification. MS-BiLSTM leverages speaker role information to learn speaker-specific context for each utterance. This contextual representation is concatenated with the utterance embedding (C2) and passed through another MS-BiLSTM (C3) which focuses on fine-grained categorization. Both tasks are jointly learned. Refer to Section \ref{sec:utt_class} for more details.}
\end{figure*}
In this paper, we propose a novel modeling approach that embodies insights regarding the organization of task-oriented conversations in order to improve performance at utterance classification over SOTA baseline approaches.  Task-oriented conversations involve sharing task-relevant information that may be useful as the task ensues~\cite{topic_summ1,topic_summ2}. Unfortunately, human-to-human conversations are less well structured than expository text, which is more often the source material for information extraction and summarization.  
Expository text is typically structured top-down and organized around information flow. Task-oriented conversations, on the other hand, are typically organized around the task and knowledge of task structure provides an implicit scaffold for understanding. 
Thus speakers feel free to elide or imply important information rather than making it explicit.  The challenges have been well documented~\cite{waitzkin1989critical,lacson2006automatic}. Prior work in utterance classification is a source of 
SOTA modeling approaches that perform relatively well despite these challenges while leaving much room for improvement.

Our evaluation in this paper specifically focuses on doctor-patient interactions.  Doctor-patient interactions are task-oriented, expert-layperson interactions in which the concerns voiced by the layperson (e.g., symptoms), the underlying issue identified by the expert (e.g., complaint) and the prescribed solutions (e.g., medications) play a crucial part. Customer-service chats are another example of such dialogue. As in the general case, topic switching abounds: the doctor may jump from a question about a symptom to a statement providing initial assessment then back again, with or without waiting for a reply from the patient (which may, itself, be responsive or introduce a new concern). In addition, the participants make unequal contributions to different parts of the schema due to the inherent asymmetry between their roles in terms of knowledge and authority. Despite these challenges, humans are able to communicate very effectively in this way.  Because of that, the issues increase as the conversation progresses and more shared context is built up, in part because of a certain amount of shared domain knowledge, despite differences in the extent and phrasing of it. 
In response to these insights, our proposed model, which we refer to as \method{}, 
integrates elements of discourse structure and ontological knowledge to improve utterance classification, the impact of which is also observed in a downstream extraction task.  We evaluate the approach on a corpus of nearly 7,000 doctor-patient interactions as a case study.

Our proposed method, \method{}\footnote{\url{https://github.com/sopankhosla/MedFilter}}, is illustrated in Figure~\ref{fig:workflow} and described in detail in Section~\ref{sec:utt_class}.  Its architecture specifically reflects an awareness of the challenges above and begins to address them. In particular, the speaker's role (i.e., doctor, patient, and other) and position within the interaction are both introduced as structuring variables.  Insights from ontological knowledge are also made available through a domain ontology: specifically, the Unified Medical Language System (UMLS)~\cite{umls}.
From a more technical perspective, the architecture introduces a novel Multi-Speaker BiLSTM to learn role-specific context representations. \method{} also benefits from the incorporation of a hierarchical loss that jointly learns the coarse-grained task of predicting medical relevance to improve fine-grained topic-based utterance classification.
The ability to extract medically relevant utterances from doctor-patient conversations and categorize them into the medical topics/categories has a substantial practical impact in medical practice \cite{naacl18_med_scribe,med_scribe_challenge}.


\begin{figure}[t]
	\centering
	\includegraphics[width=0.9\linewidth]{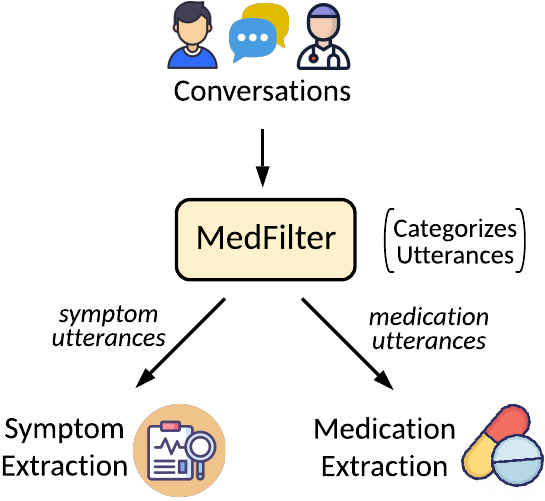}
	\captionsetup{font=small,labelfont=small}
	\caption{\label{fig:main_fig}\method{} as a part of extraction pipeline.}
\end{figure}

	\section{Related Work}
\label{sec:related_work}
\noindent\textbf{Dialogue Summarization:}
In addition to the challenges noted earlier in the paper, other linguistic phenomena such as backchannels, false starts, and topic diffusion are prominent in human-to-human conversations.  They add noise, which challenges the capabilities of otherwise effective sumarization approaches such as  pointer-generator networks~\cite{see2017get,liu2019topicsumm}.

Some prior work has relied on an Information Extraction (IE) based approach to extract details about individual medical entities such as symptoms or medications~\cite{acl19_symptom,diagnosis_detection}.  
However, recently, multiple studies~\cite{lacson2006automatic,kocaballi2019network,topic_summ1,liu2019topicsumm,park2019detecting} have shown the benefits of using the topical structure in goal-oriented dialogues to improve summarization.  Within that scope, \newcite{topic_summ1} introduce \textit{key-point sequences} that describe the logical topic flow of the summary of customer-service chats. 
They propose a hierarchical transformer to predict these topics (key-points) for each utterance and use them as auxiliary labels to guide the summarization. 

This past work inspires our work in which we extend the approach and then apply it in the more challenging domain of doctor-patient interactions.  We consider it more challenging both in terms of the number of utterances per conversation (avg. 225 vs 20) and topic switches~\cite{kocaballi2019network}.  To improve the key-point sequence utterance-level topic classification approach \cite{topic_summ1}, we propose \method{} that models speaker-specific context augmented with ontological knowledge and a hierarchical loss function. 
\noindent\textbf{Intent Classification:} The problem of classifying utterances into medical topics/categories has many similarities with the task of utterance-level intent classification~\cite{intent_1,intent_2,intent_context}. In our case, medical categories act as coarse-grained intents that drive the content of the discussion. 
Much of the previous work in intent classification caters to creating better dialog agents that condition their responses on the intent of the previous utterance~\cite{multiwoz,rasa}. For instance,~\citet{intent_3,intent_4} propose intent classification as a text classification task where each utterance is considered a complete, independent command. However, this is not true in our case as the discussion about a medical category might range over multiple utterances, each dependent on context. Hence, we tackle the classification problem as a sequence-labeling task.

\noindent\textbf{Sequence Labeling in Dialogue:}
Most prior work that employs sequence labeling for utterance classification in dialogues \cite{da_model1,da_model2,aghmn} evaluates their systems on dialogue-act classification \cite{mrda,swda2} or emotion recognition datasets \cite{meld}. In this paper, we adopt state-of-the-art modeling approaches from the emotion recognition task~\cite{aghmn,higru-sf} to serve as baselines in our evaluation since our task has not previously been benchmarked. 
	\section{Proposed Method: \method{}}
\label{sec:utt_class}

The overall architecture of \method{} is shown in Figure \ref{fig:workflow}.\
The input to \method{} is a transcribed clinical conversation $\m{C}$ of form $\{ u_1, u_2, ..., u_n\}$, where each $u_i$ represents an utterance.
Each utterance in the conversation is passed through a BERT-based encoder (Fig.~\ref{fig:workflow}A and Sec.~\ref{sec:bert_encode}) to get a fixed-dimensional representation. Contextual information such as speaker role, the utterance's position in the conversation, and ontological knowledge
(Fig. \ref{fig:workflow}B and Sec.~\ref{sec:context_info}) is then appended to the BERT representation. The encoding is input to the coarse Multi-Speaker BiLSTM (MS-BiLSTM) model (Fig. \ref{fig:workflow}C1) followed by a fully-connected layer to classify the relevance of utterances for topical classification. The representation created by MS-BiSLTM (Coarse) is then concatenated with the utterance encoding (Fig. \ref{fig:workflow}C2) and the resulting vector is fed to the fine-grained MS-BiLSTM (Fig. \ref{fig:workflow}C3) to classify utterances into different medical categories (Sec.~\ref{sec:utt_pred}). \method{} is jointly optimized on both classification tasks.
\subsection{BERT-based Encoder} 
\label{sec:bert_encode}
Given the superior modeling capabilities of long-range dependencies in Transformer-based models \cite{transformer}, we use pre-trained BERT \cite{bert} for encoding each utterance $u_i$. We first encode each token in the utterance using BERT, i.e., $[\bm{h}^{\mathtt{BERT}}_{i1}, \bm{h}^{\mathtt{BERT}}_{i2}, ..., \bm{h}^{\mathtt{BERT}}_{im}]$, where $\bm{h}_{ij}$ represents BERT-encoding of $j^{th}$ token of $u_i$. Now, following \newcite{sbert}, we use \texttt{MEAN} pooling for obtaining a representation for the entire utterance ($\bm{h}_i^{\mathtt{Text}}$).
Since the original pre-trained BERT model is trained on a general web corpus such as Wikipedia, it might not generalize well to our corpus. Therefore, 
we further fine tune the BERT model in a supervised manner for the task of predicting the utterance type. 
\subsection{Contextual Information}
\label{sec:context_info}
In addition to encoding the text of an utterance, we also make use of the following types of contextual information.

	\textbf{1. Speaker Role Info:} In conversations in general, speaker identity helps ground co-references like \textit{I, You}. In doctor-patient conversations, each of the speakers play a 
	specific role in the goals of the interaction. For example, the doctor is more likely to discuss medications than the patient. To allow the representation to be sensitive to speaker information, we map the speaker roles, namely, doctor, patient, and other, to a $d$-dimensional embedding ($\bm{h}_i^{\mathtt{speaker}}$) which is learned during training and given to the model along with the text-based representation. 
	
	\textbf{2. Positional Info:}
	Clinical conversations often follow a pattern where topics like symptoms and complaints are discussed earlier in the dialog and prescribed medications are narrated in the middle or toward the end.
	To include this signal in \method{}, we partition all the utterances in a conversation into $k$ equal parts based on their position. For instance, if the conversation has $40$ utterances and $k=4$ then the initial $10$ belong to $1^{st}$ partition and the next $10$ belong to $2^{nd}$ and so on. Similar to speaker role information, a trainable embedding is associated with each partition ($\bm{h}_i^{\mathtt{position}}$).
	
    \textbf{3. Ontological Knowledge:} UMLS (Unified Medical Language System) \cite{umls} is a combination of a semantic network and a meta-thesaurus. The semantic network consists of a set of 127 broad subject categories, or semantic types, which provide a consistent categorization of all concepts represented in the meta-thesaurus. In \method{}, we use Quick-UMLS \cite{qumls}, which identifies clinical mentions in an utterance and retrieves the associated UMLS Concept Unique Identifers (CUIs) and semantic type, to inform our model about the type of medical phrases present in the input. We believe that types such as \textit{Pharmacologic Substance}, \textit{Symptoms}, and \textit{Diseases} can be helpful in correctly classifying the utterances. We assign a trainable embedding to each semantic type. However, since each utterance can contain multiple clinical mentions of varied semantic types, we average the semantic-type embeddings for each mention present in the utterance and pass it to the model ($\bm{h}_i^{\mathtt{semantic}}$).
\subsection{Utterance Prediction}
\label{sec:utt_pred}
The classifier takes in the 
extended representation for each utterance $u_i$ in the conversation given as
\begin{equation*}
    \bm{h}_{i}  =  [\bm{h}_i^{\mathtt{Text}}; \bm{h}_i^{\mathtt{speaker}}; \bm{h}_i^{\mathtt{position}}; \bm{h}_i^{\mathtt{semantic}}].
\end{equation*}
To explicitly model the separate roles performed by each speaker (as discussed in Section~\ref{sec:intro2}), we propose a novel module \textbf{Multi-Speaker BiLSTM (MS-BiLSTM)} that includes speaker-level BiLSTMs to learn the context for each speaker type separately. We note, for example, that when the doctor is prescribing medications to the patient, she is more likely to expand on her previous utterance in order to discuss different details about the medicine, whereas the patient is most likely to give simple acknowledgments or ask questions in her turn.
Having separate speaker-level BiLSTMs allows MS-BiLSTM to model this difference in the use of context.

MS-BiLSTM
takes in $\bm{h}_i$ and $s_i$ (utterance's speaker) as input. $\bm{h}_i$ is passed through a background BiLSTM ($\mathtt{BiLSTM_{bg}}$) and different speaker-level BiLSTMs ($\mathtt{BiLSTM_{s}}$). Thus, if there are 3 speaker roles in the conversation, then the extended representation for each utterance ($\bm{h}_{i}$) would be input to 4 BiLSTMs (1 background BiLSTM + 3 speaker BiLSTMs). The hidden representations from $\mathtt{BiLSTM_{bg}}$ and $\mathtt{BiLSTM_{s_i}}$ are combined using a sigmoid gate that is learned during training:
\begin{equation*}
    \begin{split}
        \hat{\bm{h}}_i^{\mathtt{bg}} & = \mathtt{BiLSTM_{bg}}(\bm{h}_{i}) \\
        \hat{\bm{h}}_i^{\mathtt{s_{j}}} & = \mathtt{BiLSTM_{s_{j}}}(\bm{h}_{i}), \hspace*{0.1cm} \forall j \in \text{speakers} \\
        g^{s} & = \sigma(w_g) \\
        \bm{h}'_i & = g^{s_i} * \hat{\bm{h}}_i^{\mathtt{s_i}} + (1 - g^{s_i}) * \hat{\bm{h}}_i^{\mathtt{bg}}, \\
        \bm{h}'_i & = \mathtt{MS}\text{-}\mathtt{BiLSTM}(\bm{h}_{i}, s_i) .
    \end{split}
\end{equation*}%
Each speaker-level BiLSTM ($\mathtt{BiLSTM_{s_i}}$) only receives gradients for that speaker's utterance ($u_i$) thus focusing on role-specific context. The gate between $\hat{\bm{h}}_i^{\mathtt{s_i}}$ and $\hat{\bm{h}}_i^{\mathtt{bg}}$ controls the relative importance of the role-specific and general-context representation learned by speaker-level and background BiLSTMs respectively.

\vspace{3mm}
\noindent In this paper, we focus on classifying an utterance into one or more out of three categories, namely \textit{symptoms}, \textit{complaints}, and \textit{medications}.
However, these categories can be combined together to create a coarse-grained task of predicting if the utterances are \textit{medically relevant}. 
We leverage this coarse-grained supervision to create a hierarchical model with a joint-learning loss. 

\vspace{3mm}
\noindent\textbf{Hierarchical Modeling:} In this architecture, the extended representation ($\bm{h}_i$) and the corresponding speaker role ($s_i$) are first passed through a coarse-grained MS-BiLSTM and a fully-connected layer followed by softmax to be classified into one of the two categories \{\textit{Medically Relevant}, \textit{Irrelevant}\}. The representation $\bm{h}_i^{\mathtt{coarse}}$ learned by this MS-BiLSTM would model the differences between medically relevant and irrelevant text which can also benefit fine-grained classification. Hence, $\bm{h}_i^{\mathtt{coarse}}$ is concatenated with $\bm{h}_i$ and sent to the fine-grained MS-BiLSTM which focuses on the multi-label classification into the three categories discussed earlier:\vspace{-0.0cm}%
\begin{equation*}
    \begin{split}
        \bm{h}_i^{\mathtt{coarse}} &= \mathtt{MS}\text{-}\mathtt{BiLSTM_{coarse}}(\bm{h}_{i}, s_i), \\
        \bm{h}_{i}'' &= [\bm{h}_i; \bm{h}_i^{\mathtt{coarse}}], \\
        \bm{h}_i^{\mathtt{fine}} &= \mathtt{MS}\text{-}\mathtt{BiLSTM_{fine}}(\bm{h}''_{i}, s_i), \\
        \bm{p}_{\mathtt{coarse}} &= softmax(\bm{W}_c \bm{h}_i^{\mathtt{coarse}} + \bm{b}), \\
        \bm{p}_{\mathtt{fine}} &= \sigma(\bm{W}_f \bm{h}_i^{\mathtt{fine}} + \bm{b}). \\
    \end{split}
    \vspace{-0.0cm}
\end{equation*}%
Both tasks are jointly optimized
and the hyperparameter $\beta$ controls the relative strength of the medical-relevance classification loss ($\bm{L}_{\mathtt{coarse}}$):\vspace{-0.0cm}%
\begin{equation*}
\bm{L} = \bm{L}_{\mathtt{fine}} + \beta\bm{L}_{\mathtt{coarse}}.
\end{equation*}%
Such a loss function could also be used in other utterance classification tasks where classes follow a hierarchical structure. For instance, in emotion classification \cite{meld}, the fine-grained categories (e,g, happiness, anger, etc.) can be combined to create an \textit{emotive} class, and a coarse-grained classifier could be used to learn features that differentiate between \textit{emotive} and \textit{neutral} utterances.

\vspace{3mm}
	\section{Experimental Setup}
\label{sec:setup}
\subsection{Corpus Description}
\label{sec:data}
Our data set comprises 6,862 annotated transcripts of real and de-identified doctor-patient conversations with an average of 225 utterances per conversation, primarily from the doctor and patient but occasionally including contributions from nurses, caregivers, and other attendees as well.
The annotation guidelines were developed by a team of professional medical scribes and NLP experts. Annotators were trained to identify the medically-relevant utterances in a given conversation and assign one or more (out of 15 possible) tags to each utterance. Each of these tags represents a medical category like \textit{symptom}, \textit{previous medical history}, \textit{diagnosis} etc. Most conversations contain some informal, social interactions with utterances that are irrelevant to the downstream clinical tasks.\footnote{An example dialogue is included in Appendix (Sec. ~\ref{app:ex_conv}).}

In this work, we leverage the labels to train \method{} on the task of utterance classification and focus on three categories, namely, \textit{symptoms}, \textit{complaints}, and \textit{medications}, where \textit{medications} include past/current medications taken by the patient and prescriptions given by the doctor.\footnote{Additional statistics are included in Appendix (Sec. ~\ref{app:corpus_stats}).} We choose the above-mentioned categories as they are found in every office visit, and most closely generalize to other domains like customer-service chats.
However, our approach can be easily generalized for capturing other aspects such as \textit{previous medical history}, \textit{diagnosis}, and \textit{assessments} as well. 
We set aside a random sample of 627 and 592 conversations for validation and testing respectively.
\subsection{Baselines}
\label{sec:baselines}
Since sequence-labelling models haven't been applied to utterance classification in doctor-patient conversations previously, we compare our proposed method, \method{}, against baseline methods that give SOTA results on utterance-level emotion recognition data sets. \textbf{HiGRU-sf} \cite{higru-sf} is a hierarchical gated recurrent unit (HiGRU) framework with an utterance-level GRU and a conversation-level GRU. 
\textbf{BiF-AGRU} \cite{aghmn} denotes a two-level BiGRU fusion model with uni-directional AGRU for attentive context representation.
\textbf{UniF-BiAGRU} is similar to BiF-AGRU, but
uses a uni-directional GRU for contextual utterance representation and a bi-directional AGRU for attentive context. For implementation, we use the official code provided by the authors.\footnote{\url{https://github.com/wxjiao/HiGRUs}}\footnote{\url{https://github.com/wxjiao/AGHMN}}







\vspace{1.5mm}
\noindent\textbf{Evaluation Metric:}
We use the mean area under the PR curve (AUC), a widely used metric in multi-label classification setting \cite{auc1,auc2}, as our evaluation metric.
It is also used for early stopping and hyperparameter tuning.\footnote{Hyperparameters are included in Appendix (Table~\ref{tab:hp1})} 

	\section{Utterance Classification Results}
\label{sec:results}
\method{} performs better than any of the baseline approaches in assigning utterances in doctor-patient conversations to medically relevant categories. 
Table~\ref{tab:embedding} presents the AUC scores for different utterance-labeling models on our test set. Each result is the mean of 5 independent runs with different seeds. 
\begin{table}[t]
	\centering
	\resizebox{\linewidth}{!}{
	\begin{tabular}{clc}
		\toprule
		& \textbf{Methods} & \textbf{AUC} (x100) \\
		\midrule
		\multirow{3}{*}{\textbf{Baselines}} & UniF-BiAGRU & 40.9 (0.51) \\
		& BiF-AGRU & 40.9 (0.37) \\
		& HiGRU-sf & 43.1 (0.45) \\
		\midrule
        \multirow{3}{*}{\textbf{BERT variants}} & BERT & 33.5 (0.08) \\
		& Clinical BioBERT-FT & 36.1 (0.11) \\
		& BERT-FT & 36.2 (0.08) \\
        \midrule
		\multirow{2}{*}{\textbf{With Context}}& BERT BiLSTM FT & 44.5 (0.22) \\
		& BERT-FT BiLSTM & 45.8 (0.16)\\
		\midrule
		\multirow{1}{*}{\textbf{Our Method}} & \method{} & \textbf{47.2} (0.26)\\
		\bottomrule
	\end{tabular}
	}
	\captionsetup{font=small,labelfont=small}
	\caption{Utterance classification results on the test-set \textbf{(Avg. (std. dev.))}. Results on valid-set are shown in the Appendix. The improvements are statistically significant (p < 0.01).}
	\label{tab:embedding}
	\vspace{-0.3cm}
\end{table}

A BERT-based classifier that passes the mean of token-level embeddings through an FC layer gives a low score of $33.5$ AUC. When the BERT encoder is fine-tuned along with the classification layer (BERT-FT), the performance jumps to $36.2$ underlining the benefits of fine-tuning BERT~\cite{bert}. We also find that using Clinical BioBERT-FT (fine-tuned) does not beat BERT-FT. 
This is partly because the former is further pre-trained on MIMIC notes~\cite{clinbert} which are much more formal than medical conversations and thus the additional knowledge does not transfer well to our corpus. 

Adding context to BERT-based models
, using, e.g. BiLSTM, 
gives substantial boosts.  End-to-end fine-tuned BERT BiLSTM (BERT BiLSTM FT) performs worse than BERT-FT BiLSTM that passes fine-tuned BERT embeddings through a BiLSTM as non-learnable features.
\method{}, which further includes contextual information, uses MS-BiLSTM in place of BiLSTM, and optimizes a hierarchical loss, significantly outperforms all baselines and obtains $1.4$ absolute AUC points over BERT-FT BiLSTM ($2^{nd}$ best). It also surpasses emotion recognition SOTA methods like HiGRU-sf by $4.1$ AUC points.
\\

\begin{figure}[t]
	\centering
	\includegraphics[width=\linewidth]{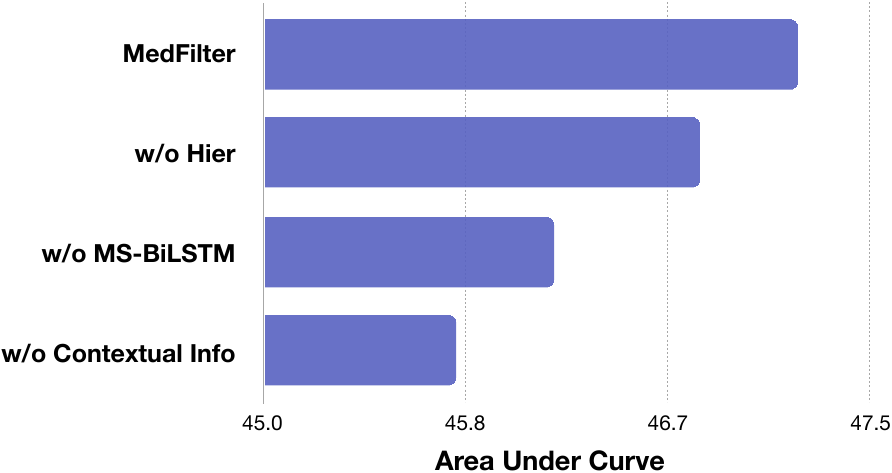}
	\captionsetup{font=small,labelfont=small}
	\captionof{figure}{\label{fig:ablation} Cumulative Ablation Results}
\end{figure}

\vspace{-3mm}
\noindent\textbf{Ablation Results:}
\label{sec:ablation_results}
To understand the importance of each module in \method{}, we perform a cumulative ablation study (Figure~\ref{fig:ablation}). We find that removing individual modules results in notably reduced performance.
The model that does not incorporate hierarchical modeling, shows a dip of $0.4$ AUC points. This suggests that the information learned in the medical-relevance prediction layer aids the final classification task. Further, replacing MS-BiLSTM with a simple BiLSTM leads to a drop of an additional $0.6$ AUC points, revealing the importance of modeling speaker-specific context. Without contextual information, we see a reduction of $0.4$ AUC points. This shows that features like speaker role, position, and semantic types are essential for our task.

	\section{Impact of Utterance Classification on Downstream Medical Extraction}
\label{sec:footprint_extractor}
The results in the previous section portray the effectiveness of \method{} at sorting important utterances in clinical conversations into medically relevant categories. Such filtering, when included in the pipeline (for example, as a pre-processing step), can assist downstream medical processing methods to focus on utterances that contain information pertinent for their tasks (Figure~\ref{fig:main_fig}), by improving the signal-to-noise ratio in the input.
In this section, we evaluate whether the use of \method{} to prune irrelevant utterances is advantageous for symptom, medication, and complaint extraction.
 
\subsection{Task Setup}
\label{sec:task_setup}
The extractor takes the conversation as input and outputs the discussed symptoms/medications/complaints within.

\noindent Conversation-level labels for all three extraction tasks are taken from a predefined set provided by the corpus annotators. 
For \textbf{symptoms}, they
include $14$ coarse-grained classes to represent different body systems (e.g., cardiovascular) and $178$ fine-grained ones for the corresponding issues (e.g., palpitations).
Given the small size of the training data, we use the coarse-grained body-systems for symptom extraction. We then manually curate a list of different symptoms corresponding to each body-system using UMLS and use their UMLS CUIs as labels.\footnote{Refer to Table~\ref{tab:sym_labels} for the final list of symptom labels.} 

For \textbf{medications}, we manually link medication labels to their corresponding UMLS~\cite{umls} concepts and group them using hierarchies from 
NCI Thesaurus \cite{nci}.
\footnote{\url{https://ncit.nci.nih.gov}}
We pass each medication name through QuickUMLS to get a list of possible CUIs for the term in UMLS. We take
the candidate CUI with a similarity of 1 and find its NCI hierarchy in the UMLS metatheasurus. The four
topmost nodes in the hierarchy are extracted, which act as the pseudo-label for that CUI. In order to reduce
the class-imbalance, some of these hierarchies are combined to form a coarser label. This reduces the number of labels to 31. Finally, \textit{Others} label is added, which inhabits medicine names (in
the test-set) that do not correspond to any of the
previous 31 labels. This reduces the label count to $32$ for medications.\footnote{Refer to Table \ref{tab:med_labels} for the final list of medication labels.} 

\textbf{Complaints} in our corpus range from follow-up visits to disease names to vaccine requests. Similar to medication extraction, we leverage SNOMED-CT hierarchies\footnote{\url{https://www.nlm.nih.gov/healthit/snomedct/index.html}} to constraint the tag list to 11, where the first 10 represent diseases of different body systems and \textit{Others} encompasses complaints like follow-up, vaccine requests, medication refill requests, etc. (Table~\ref{tab:comp_labels}).

We use the same train/val/test split as defined for the utterance classification experiments in Section~\ref{sec:data}. The performance of the extraction pipeline is evaluated on Micro and Macro-F1 scores.

\begin{table}[t]
	\centering
	\resizebox{1.0\linewidth}{!}{
	\begin{tabular}{lcc}
		\hline
		\toprule
		\textbf{Medication Extraction} & \textbf{Macro F1} & \textbf{Micro F1}\\ 
		\midrule
		QuickUMLS (All Text) & 25.4 & 33.5\\
		\midrule
		QuickUMLS (MR BERT-FT BiLSTM) &32.6 &61.9 \\
		QuickUMLS (MU BERT-FT BiLSTM) &34.0 &67.3 \\
		\midrule
		QuickUMLS (MR \method{}) & 34.2 &62.8\\
		QuickUMLS (MU \method{}) & \textbf{35.9} &\textbf{68.9}\\
		\bottomrule
	\end{tabular}
	}
	\captionsetup{font=small,labelfont=small}
	\caption{Results(\%) for Medication Extraction. \textbf{MR}=Medically Relevant (Symptom + Complaints + Medications) Utterances, \textbf{MU}=Medication Utterances.}
	\label{tab:medication}
	\vspace{-0.27cm}
\end{table}

\subsection{Extractor Details}
All three extraction tasks are modeled as multi-label classification. We leverage a state-of-the-art medical entity-linking tool, QuickUMLS~\cite{qumls}\footnote{\url{https://github.com/Georgetown-IR-Lab/QuickUMLS}}, that takes in a conversation and outputs UMLS CUIs corresponding to all identified candidate concepts. Concepts with a similarity measure of $1$ are chosen as predictions. For symptom extraction, the predictions are compared against a manually created list of CUIs (presented in Appendix Table~\ref{tab:sym_targ}) for symptoms associated with each of the 14 Body Systems. The presence of a symptom of body-system $b$ is determined by the presence of the predicted CUIs in the target list for that body system. We compare the NCI and the SNOMED-CT hierarchies of the predicted concepts against the label hierarchies for medications and complaints, respectively. Concepts that do not fit into one of the specific categories are grouped under the label \textit{Others}. In the next section, we report the results for the best performing filtering thresholds.\footnote{Micro F1 vs filtering threshold graphs are presented in the Appendix (Figures~\ref{fig:sym_ex} and \ref{fig:med_ex}).}

\subsection{Results}
\label{sec:downstream}
We find that the performance of the baseline medication and symptom extractor QuickUMLS (All Text) is substantially boosted by filtering out irrelevant utterances (Tables~\ref{tab:medication} and \ref{tab:symptom}).
Pruning medically irrelevant utterances using \method{} (MR \method{}) improves Micro F1 by $29.3$ and $4.7$ points for medication and symptom extraction, respectively. If only the medication/symptomatic utterances (MU/SU) are input to the extractors, the results improve further. 

Results for complaint extraction are shown in Table~\ref{tab:comp}. We find that the QuickUMLS extractor does not perform well on complaint extraction. However, consistent with the other two categories' trends, pruning irrelevant utterances before sending the conversation through the extractor improves performance. Micro-F1 score increases from 35.6 for All Text to 43.7 for CU \method{}.

\begin{table}[t]
    \centering
    \resizebox{1.0\linewidth}{!}{
    \begin{tabular}{lcc}
    \hline
    	\toprule
        \textbf{Symptom Extraction} &\textbf{Macro F1} &\textbf{Micro F1}\\
        \midrule
        QuickUMLS (All Text) & 33.9 & 42.7\\
        \midrule
        QuickUMLS (MR BERT-FT BiLSTM) & \textbf{36.4} & 47.4\\
        QuickUMLS (SU BERT-FT BiLSTM) & 35.9 & 49.2\\
        \midrule
        QuickUMLS (MR \method{}) & 35.2 & 47.4\\
        QuickUMLS (SU \method{}) & 36.1 & \textbf{49.3}\\
        \bottomrule
    \end{tabular}
    }
    \captionsetup{font=small,labelfont=small}
    \caption{Results(\%) for Symptom Extraction. \textbf{MR}=Medically Relevant, \textbf{SU}=Symptom Utterances.}
    \label{tab:symptom}
\end{table}
\begin{table}[h]
    \centering
    \resizebox{1\linewidth}{!}{
    \begin{tabular}{lcc}
    \hline
    	\toprule
        \textbf{Complaint Extraction} &\textbf{Macro F1} & \textbf{Micro F1}\\
        \midrule
        QuickUMLS (All Text) & 10.0 & 35.6\\
        \midrule
        QuickUMLS (MR BERT-FT BiLSTM) & 10.9 & 40.3\\
        QuickUMLS (CU BERT-FT BiLSTM) & 11.1 & 43.0\\
        \midrule
        QuickUMLS (MR \method{}) & 11.1 & 40.7\\
        QuickUMLS (CU \method{}) & \textbf{11.1} & \textbf{43.7}\\
        \bottomrule
    \end{tabular}
    }
    \captionsetup{font=small,labelfont=small}
    \caption{Results(\%) for Complaint Extraction (CE). \textbf{MR}=Medically Relevant, \textbf{CU}=Complaint Utterances.}
    \label{tab:comp}
\end{table}
\begin{figure*}[t]
	\centering
	\begin{subfigure}[b]{0.46\linewidth}
		\centering
		\includegraphics[width=0.84\textwidth]{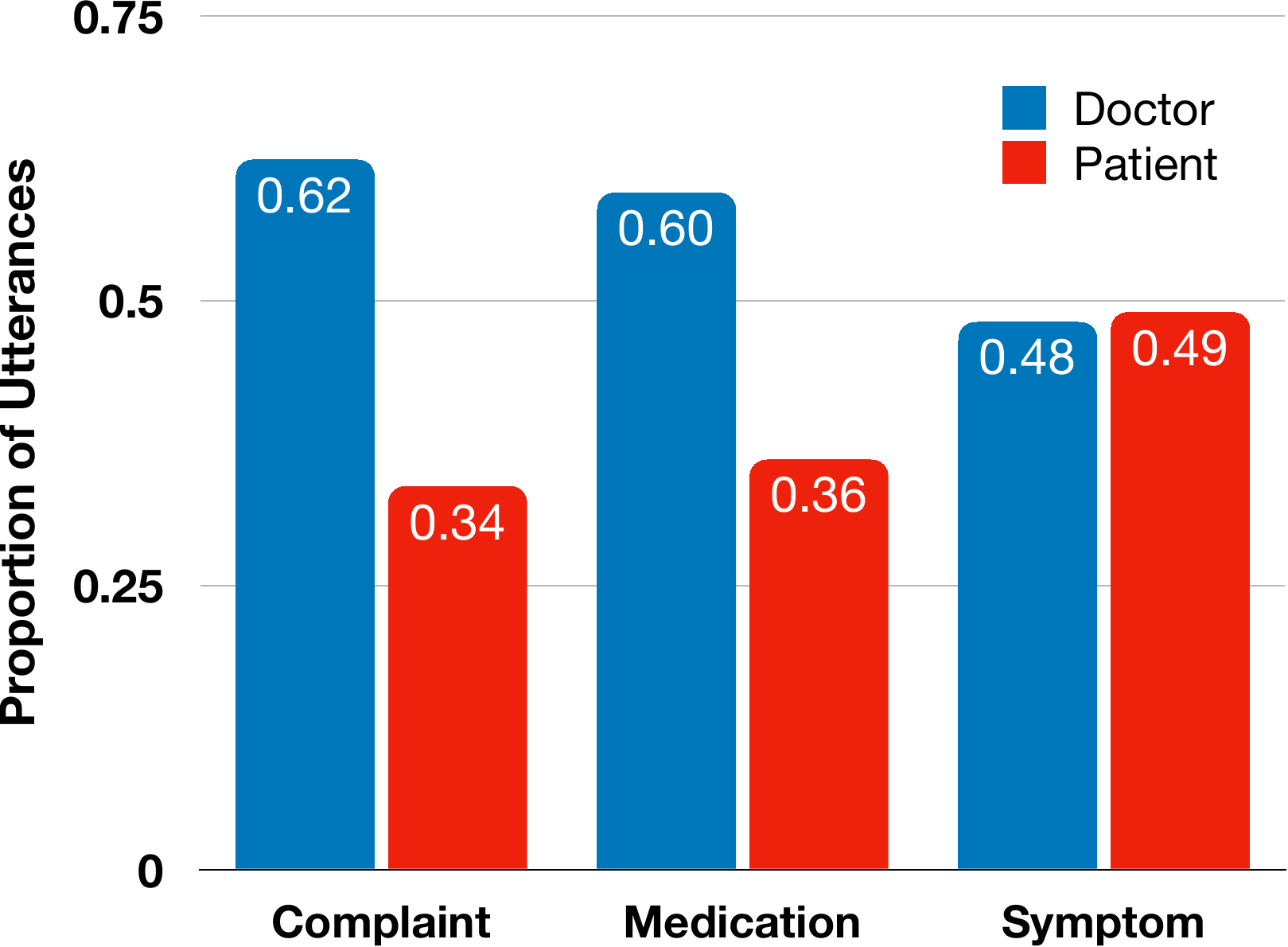}
		\subcaption{Proportion of utterances, in each medical category, spoken by different speaker roles.}
		\label{fig:speaker_side}
	\end{subfigure}
	\hspace{0.5cm}
	\begin{subfigure}[b]{0.46\linewidth}
		\centering
		\includegraphics[width=0.9\textwidth]{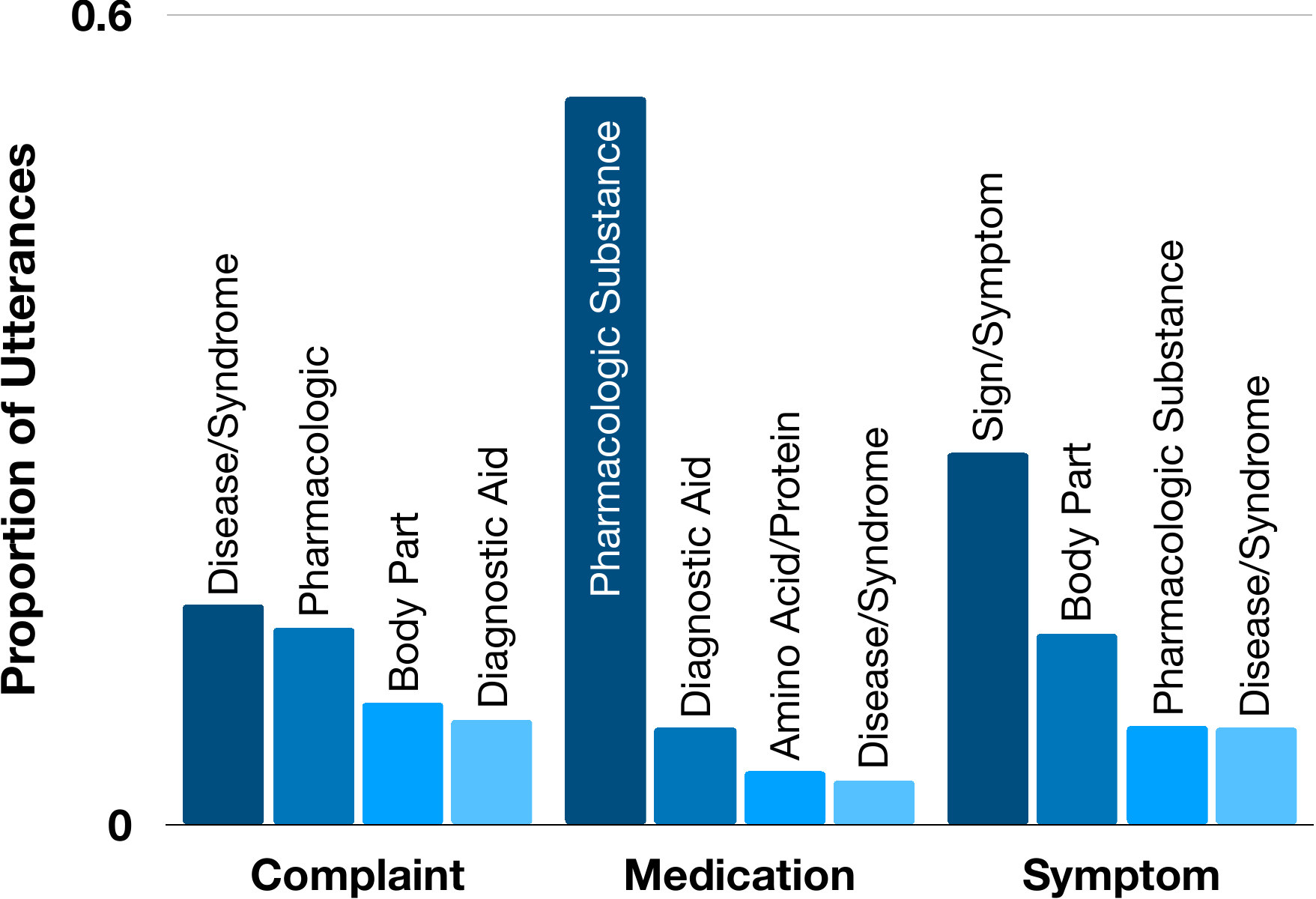}
		\subcaption{Most frequent semantic types in terms of the proportion of utterances they occur in.}
		\label{fig:semantic_side}
	\end{subfigure}
	\captionsetup{font=small,labelfont=small}
	\caption{Contextual Information: Different speaker roles contribute asymmetrically towards different medical topics/categories in the dialogue~(Figure \ref{fig:speaker_side}). 
    Furthermore, phrases with UMLS semantic types \textit{Pharmacologic Substance}, \textit{Sign/Symptom}, and \textit{Disease/Syndrome} occur quite frequently in medical, symptom, and complaint utterances respectively (Figure \ref{fig:semantic_side}).
	}
	\label{fig:side}
\end{figure*}
Pruning done using \method{} seems to be more beneficial than BERT-FT BiLSTM ($2^{nd}$ best utterance classifier in Table~\ref{tab:embedding}) for medication and complaint extraction, however they perform equally well for symptom extraction.
This suggests that the benefits from the inclusion of discourse structure, domain knowledge, and a hierarchical loss function, do not transfer well to symptom extraction. In Section~\ref{sec:error}, we investigate the kinds of utterance classification errors \method{} makes, that need to be addressed to further improve the symptom extraction pipeline.




	\section{Discussion}
\label{sec:discussion}

\noindent\textbf{Why does contextual information help?}
Ablation results (Figure~\ref{fig:ablation}) show that incorporating speaker role information and UMLS semantic-type information provides significant improvements in AUC scores for utterance classification. In Figure~\ref{fig:speaker_side}, we plot the proportion of utterances from different medical categories against their speakers. While both parties contribute equally to symptom discussions, there is a clear asymmetry in the number of \textit{medication} and \textit{complaint} utterances spoken by the doctor and the patient, explaining 
the contribution of speaker role information in differentiating \textit{medication/complaint} utterances from others. 

We also plot the distribution of the four most frequent UMLS semantic types present in the utterances of different medical categories (Figure~\ref{fig:semantic_side}). For medications, we find that UMLS entities with semantic type \textit{Pharmacologic Substance} are present in more than $55\%$ of the medication utterances indicating that its detection is a knowledge-dependent task. Similarly, and supporting our hypothesis, \textit{Disease/Syndrome} and \textit{Sign/Symptom} are the most frequent semantic types in complaint and symptom utterances, respectively. \\

\noindent\textbf{Error Analysis:}
\label{sec:error}
In this section, we present a deeper analysis of some of the systematic knowledge-extraction errors made by \method{} that limit its performance in recognizing medically-relevant utterances. 

    \textbf{1. Informal Language:} The model sometimes overlooks informal references to symptoms. For instance, utterances such as \textit{PT: I feel something unusual in my leg} or \textit{PT: My heart beats funny!} discuss musculoskeletal and cardiovascular symptoms but do not use medical terms to refer to them. These patterns seem to be more frequent in patient utterances, likely because they are less familiar with medical terminology.
    Off-the-shelf entity-linkers, like QuickUMLS~\cite{qumls}, do not transfer well to spoken medical conversations. They are unable to recognise the correct UMLS concepts (and semantic types) corresponding to the colloquial symptomatic phrases which reduces their effectiveness as features.\footnote{\textit{Sign/Symptom} entities are identified in less than $30\%$ of total symptomatic utterances (Figure~\ref{fig:semantic_side}).}
    For instance, for the utterance \textit{PT: My heart is racing.}, QuickUMLS outputs \ref{eq:1}A rather than \ref{eq:1}B:
	\vspace{-3mm}
	
    {\small
    \begin{equation}
    \begin{split}
        &\text{\textbf{Input:} PT: My heart is racing} \\
        &\text{\textbf{(A):} PT: My~} [\text{\textbf{heart}}]_{\mathtt{body\_part}} \text{~is racing.} \\
        &\text{\textbf{(B):} PT: My~} [[\text{\textbf{heart}}]_{\mathtt{body\_part}} \text{~\textbf{is racing}}]]_{\mathtt{symptom}}\text{.}
    \label{eq:1}
    \end{split}
    \end{equation}}
	\vspace{-2mm}
    \normalsize
    
    \textbf{2. Physical Manifestations of Symptoms:} Internal symptoms often manifest themselves physically as a digression from the natural ability to perform typical activities. For instance, when the patient says \textit{I can't do anything after I'm back from office} or \textit{I can only walk up one flight of stairs},
    she might be implicitly mentioning a cardiovascular symptom.
    A sizeable subset of such examples includes usage of duration or frequency to convey the implicit deviation, like 
    \vspace{-3mm}
    
    \small
    \begin{equation*}
    \begin{split}
        &\text{\textbf{Duration:} PT: I only sleep for \textbf{4 hours},} \\
        &\text{\textbf{Frequency:} PT: I go to the bathroom \textbf{10 times} at night,} \\
        &\text{\textbf{Quantity:} DR: I see you are up to \textbf{300 pounds} now.}
    \end{split}
    \end{equation*}
    \normalsize
\vspace{-0.2cm}
\section{Future Work} 
For a system to correctly classify the samples of the above two categories, it needs both to generalize to patient-generated language, and to have a semantic understanding of whether the description strays from normal. Incorporating data from online self-disclosure sites like medical subreddits and discussion forums~\cite{basaldella2019bioreddit} during training might prove beneficial for learning better representations for such vocabulary. 
Concept normalization data sets~\cite{concept_norm_1,concept_norm_2} could also be leveraged in this regard.
Our approach of training the BERT encoder separately from the context encoder would allow \method{} to learn from such non-dialogue resources. 


Extraction tasks (Section~\ref{sec:footprint_extractor}) mostly evaluate the ability of \method{} to recognize utterances that contain the most information about the name or type of the medication, symptom, or complaint. However, to quantify the context-level benefits of \method{}, especially the speaker-specific context modeling (MS-BiLSTM), on downstream processing, we need to evaluate the system on problems like regimen extraction~\cite{diagnosis_detection} or symptom summarization~\cite{liu2019topicsumm}. Such tasks require utterance classification models to correctly identify utterances that discuss fine-grained details about the topic and would therefore evaluate a model's ability to solve multiple challenges like coreference resolution, speaker-specific context detection, thread identification, etc. Such an evaluation is a part of future work.
	\vspace{-0.13cm}
\section{Conclusion}
\label{sec:conclusion}
\vspace{-0.13cm}
In this paper, we have proposed a novel text classification approach that specifically leverages insights into the organization of task-oriented conversations in order to improve performance at topic-based utterance classification over SOTA baseline approaches.  In particular, we have demonstrated that our utterance classification model, \method{}, benefits from discourse information, domain knowledge, speaker-specific context modeling, and a hierarchical loss to 
reach a new state-of-the-art performance on a doctor-patient interactions dataset.
We find that using topic-based utterance classification in general, and \method{} in particular, as a pre-processing step before medical extraction tasks, significantly improves the extraction scores. We believe that the contributions made in this work would also generalize to other kinds of expert-lay dialogue like customer-service chats.

	
	\section*{Acknowledgements}
    We thank the anonymous EMNLP reviewers for their insightful comments. We are also grateful to the members of
    the TELEDIA group at LTI, CMU for the invaluable feedback. This work was funded in part by Dow Chemical, University of Pittsburgh Medical  Center, and Microsoft.
	
	
	\bibliographystyle{acl_natbib}
	\bibliography{references}

\begin{thebibliography}{38}
\expandafter\ifx\csname natexlab\endcsname\relax\def\natexlab#1{#1}\fi

\bibitem[{Alsentzer et~al.(2019)Alsentzer, Murphy, Boag, Weng, Jin, Naumann,
  and McDermott}]{clinbert}
Emily Alsentzer, John Murphy, William Boag, Wei-Hung Weng, Di~Jin, Tristan
  Naumann, and Matthew McDermott. 2019.
\newblock \href {https://doi.org/10.18653/v1/W19-1909} {Publicly available
  clinical {BERT} embeddings}.
\newblock In \emph{Proceedings of the 2nd Clinical Natural Language Processing
  Workshop}, pages 72--78, Minneapolis, Minnesota, USA. Association for
  Computational Linguistics.

\bibitem[{Basaldella and Collier(2019)}]{basaldella2019bioreddit}
Marco Basaldella and Nigel Collier. 2019.
\newblock Bioreddit: Word embeddings for user-generated biomedical nlp.
\newblock In \emph{Proceedings of the Tenth International Workshop on Health
  Text Mining and Information Analysis (LOUHI 2019)}, pages 34--38.

\bibitem[{Bocklisch et~al.(2017)Bocklisch, Faulkner, Pawlowski, and
  Nichol}]{rasa}
Tom Bocklisch, Joey Faulkner, Nick Pawlowski, and Alan Nichol. 2017.
\newblock Rasa: Open source language understanding and dialogue management.
\newblock \emph{arXiv preprint arXiv:1712.05181}.

\bibitem[{Bodenreider(2004)}]{umls}
Olivier Bodenreider. 2004.
\newblock The unified medical language system (umls): integrating biomedical
  terminology.
\newblock \emph{Nucleic acids research}, 32 Database issue:D267--70.

\bibitem[{Budzianowski et~al.(2018{\natexlab{a}})Budzianowski, Wen, Tseng,
  Casanueva, Stefan, Osman, and Ga{\v{s}}i\'c}]{multiwoz}
Pawe{\l} Budzianowski, Tsung-Hsien Wen, Bo-Hsiang Tseng, I{\~n}igo Casanueva,
  Ultes Stefan, Ramadan Osman, and Milica Ga{\v{s}}i\'c. 2018{\natexlab{a}}.
\newblock Multiwoz - a large-scale multi-domain wizard-of-oz dataset for
  task-oriented dialogue modelling.
\newblock In \emph{Proceedings of the 2018 Conference on Empirical Methods in
  Natural Language Processing (EMNLP)}.

\bibitem[{Budzianowski et~al.(2018{\natexlab{b}})Budzianowski, Wen, Tseng,
  Casanueva, Ultes, Ramadan, and Gasic}]{intent_2}
Pawe{\l} Budzianowski, Tsung-Hsien Wen, Bo-Hsiang Tseng, I{\~n}igo Casanueva,
  Stefan Ultes, Osman Ramadan, and Milica Gasic. 2018{\natexlab{b}}.
\newblock Multiwoz-a large-scale multi-domain wizard-of-oz dataset for
  task-oriented dialogue modelling.
\newblock In \emph{Proceedings of the 2018 Conference on Empirical Methods in
  Natural Language Processing}, pages 5016--5026.

\bibitem[{Chen et~al.(2019)Chen, Fu, Hu, Zhang, Zhou, Li, and Bao}]{intent_3}
Cen Chen, Chilin Fu, Xu~Hu, Xiaolu Zhang, Jun Zhou, Xiaolong Li, and
  Forrest~Sheng Bao. 2019.
\newblock Reinforcement learning for user intent prediction in customer service
  bots.
\newblock In \emph{Proceedings of the 42nd International ACM SIGIR Conference
  on Research and Development in Information Retrieval}, pages 1265--1268.

\bibitem[{Devlin et~al.(2019)Devlin, Chang, Lee, and Toutanova}]{bert}
Jacob Devlin, Ming-Wei Chang, Kenton Lee, and Kristina Toutanova. 2019.
\newblock \href {https://doi.org/10.18653/v1/N19-1423} {{BERT}: Pre-training of
  deep bidirectional transformers for language understanding}.
\newblock In \emph{Proceedings of the 2019 Conference of the North {A}merican
  Chapter of the Association for Computational Linguistics: Human Language
  Technologies, Volume 1 (Long and Short Papers)}, pages 4171--4186,
  Minneapolis, Minnesota. Association for Computational Linguistics.

\bibitem[{Du et~al.(2019)Du, Chen, Kannan, Tran, Chen, and
  Shafran}]{acl19_symptom}
Nan Du, Kai Chen, Anjuli Kannan, Linh Tran, Yuhui Chen, and Izhak Shafran.
  2019.
\newblock \href {https://doi.org/10.18653/v1/P19-1087} {Extracting symptoms and
  their status from clinical conversations}.
\newblock In \emph{Proceedings of the 57th Annual Meeting of the Association
  for Computational Linguistics}, pages 915--925, Florence, Italy. Association
  for Computational Linguistics.

\bibitem[{Finley et~al.(2018)Finley, Edwards, Robinson, Brenndoerfer, Sadoughi,
  Fone, Axtmann, Miller, and Suendermann-Oeft}]{naacl18_med_scribe}
Gregory Finley, Erik Edwards, Amanda Robinson, Michael Brenndoerfer, Najmeh
  Sadoughi, James Fone, Nico Axtmann, Mark Miller, and David Suendermann-Oeft.
  2018.
\newblock \href {https://doi.org/10.18653/v1/N18-5003} {An automated medical
  scribe for documenting clinical encounters}.
\newblock In \emph{Proceedings of the 2018 Conference of the North {A}merican
  Chapter of the Association for Computational Linguistics: Demonstrations},
  pages 11--15, New Orleans, Louisiana. Association for Computational
  Linguistics.

\bibitem[{Jiao et~al.(2019{\natexlab{a}})Jiao, Lyu, and King}]{aghmn}
Wenxiang Jiao, Michael~R Lyu, and Irwin King. 2019{\natexlab{a}}.
\newblock Real-time emotion recognition via attention gated hierarchical memory
  network.
\newblock \emph{arXiv preprint arXiv:1911.09075}.

\bibitem[{Jiao et~al.(2019{\natexlab{b}})Jiao, Yang, King, and Lyu}]{higru-sf}
Wenxiang Jiao, Haiqin Yang, Irwin King, and Michael~R. Lyu. 2019{\natexlab{b}}.
\newblock \href {https://doi.org/10.18653/v1/N19-1037} {{H}i{GRU}:
  {H}ierarchical gated recurrent units for utterance-level emotion
  recognition}.
\newblock In \emph{Proceedings of the 2019 Conference of the North {A}merican
  Chapter of the Association for Computational Linguistics: Human Language
  Technologies, Volume 1 (Long and Short Papers)}, pages 397--406, Minneapolis,
  Minnesota. Association for Computational Linguistics.

\bibitem[{Kazi and Kahanda(2019)}]{topic_summ2}
Nazmul Kazi and Indika Kahanda. 2019.
\newblock Automatically generating psychiatric case notes from digital
  transcripts of doctor-patient conversations.
\newblock In \emph{Proceedings of the 2nd Clinical Natural Language Processing
  Workshop}, pages 140--148.

\bibitem[{Kim et~al.(2017)Kim, Lee, and Stratos}]{intent_4}
Young-Bum Kim, Sungjin Lee, and Karl Stratos. 2017.
\newblock Onenet: Joint domain, intent, slot prediction for spoken language
  understanding.
\newblock In \emph{2017 IEEE Automatic Speech Recognition and Understanding
  Workshop (ASRU)}, pages 547--553. IEEE.

\bibitem[{Kocaballi et~al.(2019)Kocaballi, Coiera, Tong, White, Quiroz,
  Rezazadegan, Willcock, and Laranjo}]{kocaballi2019network}
Ahmet~Baki Kocaballi, Enrico Coiera, Huong~Ly Tong, Sarah~J White, Juan~C
  Quiroz, Fahimeh Rezazadegan, Simon Willcock, and Liliana Laranjo. 2019.
\newblock A network model of activities in primary care consultations.
\newblock \emph{Journal of the American Medical Informatics Association},
  26(10):1074--1082.

\bibitem[{Lacson et~al.(2006)Lacson, Barzilay, and Long}]{lacson2006automatic}
Ronilda~C Lacson, Regina Barzilay, and William~J Long. 2006.
\newblock Automatic analysis of medical dialogue in the home hemodialysis
  domain: structure induction and summarization.
\newblock \emph{Journal of biomedical informatics}, 39(5):541--555.

\bibitem[{Lee et~al.(2017)Lee, Hasan, Farri, Choudhary, and
  Agrawal}]{concept_norm_2}
Kathy Lee, Sadid~A Hasan, Oladimeji Farri, Alok Choudhary, and Ankit Agrawal.
  2017.
\newblock Medical concept normalization for online user-generated texts.
\newblock In \emph{2017 IEEE International Conference on Healthcare Informatics
  (ICHI)}, pages 462--469. IEEE.

\bibitem[{Liu et~al.(2019{\natexlab{a}})Liu, Wang, Xu, Li, and
  Ye}]{topic_summ1}
Chunyi Liu, Peng Wang, Jiang Xu, Zang Li, and Jieping Ye. 2019{\natexlab{a}}.
\newblock Automatic dialogue summary generation for customer service.
\newblock In \emph{Proceedings of the 25th ACM SIGKDD International Conference
  on Knowledge Discovery \& Data Mining}, pages 1957--1965.

\bibitem[{Liu et~al.(2017)Liu, Han, Tan, and Lei}]{da_model2}
Yang Liu, Kun Han, Zhao Tan, and Yun Lei. 2017.
\newblock \href {https://doi.org/10.18653/v1/D17-1231} {Using context
  information for dialog act classification in {DNN} framework}.
\newblock In \emph{Proceedings of the 2017 Conference on Empirical Methods in
  Natural Language Processing}, pages 2170--2178, Copenhagen, Denmark.
  Association for Computational Linguistics.

\bibitem[{Liu et~al.(2019{\natexlab{b}})Liu, Ng, Guang, Aw, and
  Chen}]{liu2019topicsumm}
Zhengyuan Liu, Angela Ng, Sheldon Lee~Shao Guang, Ai~Ti Aw, and Nancy~F. Chen.
  2019{\natexlab{b}}.
\newblock \href {https://doi.org/10.1109/ASRU46091.2019.9003764} {Topic-aware
  pointer-generator networks for summarizing spoken conversations}.
\newblock In \emph{{IEEE} Automatic Speech Recognition and Understanding
  Workshop, {ASRU} 2019, Singapore, December 14-18, 2019}, pages 814--821.
  {IEEE}.

\bibitem[{Miftahutdinov and Tutubalina(2019)}]{concept_norm_1}
Zulfat Miftahutdinov and Elena Tutubalina. 2019.
\newblock Deep neural models for medical concept normalization in
  user-generated texts.
\newblock In \emph{Proceedings of the 57th Annual Meeting of the Association
  for Computational Linguistics: Student Research Workshop}, pages 393--399.

\bibitem[{Mintz et~al.(2009)Mintz, Bills, Snow, and Jurafsky}]{auc2}
Mike Mintz, Steven Bills, Rion Snow, and Dan Jurafsky. 2009.
\newblock Distant supervision for relation extraction without labeled data.
\newblock In \emph{Proceedings of the Joint Conference of the 47th Annual
  Meeting of the ACL and the 4th International Joint Conference on Natural
  Language Processing of the AFNLP: Volume 2-Volume 2}, pages 1003--1011.
  Association for Computational Linguistics.

\bibitem[{Park et~al.(2019)Park, Kotzias, Kuo, Logan~IV, Merced, Singh, Tanana,
  Karra~Taniskidou, Lafata, Atkins et~al.}]{park2019detecting}
Jihyun Park, Dimitrios Kotzias, Patty Kuo, Robert~L Logan~IV, Kritzia Merced,
  Sameer Singh, Michael Tanana, Efi Karra~Taniskidou, Jennifer~Elston Lafata,
  David~C Atkins, et~al. 2019.
\newblock Detecting conversation topics in primary care office visits from
  transcripts of patient-provider interactions.
\newblock \emph{Journal of the American Medical Informatics Association},
  26(12):1493--1504.

\bibitem[{Poria et~al.(2019)Poria, Hazarika, Majumder, Naik, Cambria, and
  Mihalcea}]{meld}
Soujanya Poria, Devamanyu Hazarika, Navonil Majumder, Gautam Naik, Erik
  Cambria, and Rada Mihalcea. 2019.
\newblock \href {https://doi.org/10.18653/v1/P19-1050} {{MELD}: A multimodal
  multi-party dataset for emotion recognition in conversations}.
\newblock In \emph{Proceedings of the 57th Annual Meeting of the Association
  for Computational Linguistics}, pages 527--536, Florence, Italy. Association
  for Computational Linguistics.

\bibitem[{Qu et~al.(2019)Qu, Yang, Croft, Zhang, Trippas, and
  Qiu}]{intent_context}
Chen Qu, Liu Yang, W~Bruce Croft, Yongfeng Zhang, Johanne~R Trippas, and
  Minghui Qiu. 2019.
\newblock User intent prediction in information-seeking conversations.
\newblock In \emph{Proceedings of the 2019 Conference on Human Information
  Interaction and Retrieval}, pages 25--33.

\bibitem[{Quiroz et~al.(2019)Quiroz, Laranjo, Kocaballi, Berkovsky,
  Rezazadegan, and Coiera}]{med_scribe_challenge}
Juan~C. Quiroz, Liliana Laranjo, Ahmet~Baki Kocaballi, Shlomo Berkovsky, Dana
  Rezazadegan, and Enrico Coiera. 2019.
\newblock \href {https://doi.org/10.1038/s41746-019-0190-1} {Challenges of
  developing a digital scribe to reduce clinical documentation burden}.
\newblock \emph{npj Digital Medicine}, 2(1):114.

\bibitem[{Raheja and Tetreault(2019)}]{da_model1}
Vipul Raheja and Joel Tetreault. 2019.
\newblock \href {https://doi.org/10.18653/v1/N19-1373} {{D}ialogue {A}ct
  {C}lassification with {C}ontext-{A}ware {S}elf-{A}ttention}.
\newblock In \emph{Proceedings of the 2019 Conference of the North {A}merican
  Chapter of the Association for Computational Linguistics: Human Language
  Technologies, Volume 1 (Long and Short Papers)}, pages 3727--3733,
  Minneapolis, Minnesota. Association for Computational Linguistics.

\bibitem[{Reimers and Gurevych(2019)}]{sbert}
Nils Reimers and Iryna Gurevych. 2019.
\newblock \href {http://arxiv.org/abs/1908.10084} {Sentence-bert: Sentence
  embeddings using siamese bert-networks}.
\newblock In \emph{Proceedings of the 2019 Conference on Empirical Methods in
  Natural Language Processing}. Association for Computational Linguistics.

\bibitem[{Riedel et~al.(2013)Riedel, Yao, McCallum, and Marlin}]{auc1}
Sebastian Riedel, Limin Yao, Andrew McCallum, and Benjamin~M Marlin. 2013.
\newblock Relation extraction with matrix factorization and universal schemas.
\newblock In \emph{Proceedings of the 2013 Conference of the North American
  Chapter of the Association for Computational Linguistics: Human Language
  Technologies}, pages 74--84.

\bibitem[{See et~al.(2017)See, Liu, and Manning}]{see2017get}
Abigail See, Peter~J Liu, and Christopher~D Manning. 2017.
\newblock Get to the point: Summarization with pointer-generator networks.
\newblock In \emph{Proceedings of the 55th Annual Meeting of the Association
  for Computational Linguistics (Volume 1: Long Papers)}, pages 1073--1083.

\bibitem[{Selvaraj and Konam(2019)}]{diagnosis_detection}
Sai~P Selvaraj and Sandeep Konam. 2019.
\newblock Medication regimen extraction from medical conversations.
\newblock \emph{arXiv}, pages arXiv--1912.

\bibitem[{Shriberg et~al.(1998)Shriberg, Bates, Taylor, Stolcke, Jurafsky,
  Ries, Coccaro, Martin, Meteer, and Van Ess-Dykema}]{swda2}
Elizabeth Shriberg, Rebecca Bates, Paul Taylor, Andreas Stolcke, Daniel
  Jurafsky, Klaus Ries, Noah Coccaro, Rachel Martin, Marie Meteer, and Carol
  Van Ess-Dykema. 1998.
\newblock Can prosody aid the automatic classification of dialog acts in
  conversational speech?
\newblock \emph{Language and Speech}, 41(3--4):439--487.

\bibitem[{Shriberg et~al.(2004)Shriberg, Dhillon, Bhagat, Ang, and
  Carvey}]{mrda}
Elizabeth Shriberg, Raj Dhillon, Sonali Bhagat, Jeremy Ang, and Hannah Carvey.
  2004.
\newblock \href {https://www.aclweb.org/anthology/W04-2319} {The {ICSI} meeting
  recorder dialog act ({MRDA}) corpus}.
\newblock In \emph{Proceedings of the 5th {SIG}dial Workshop on Discourse and
  Dialogue at {HLT}-{NAACL} 2004}, pages 97--100, Cambridge, Massachusetts,
  USA. Association for Computational Linguistics.

\bibitem[{Sioutos et~al.(2007)Sioutos, Coronado, Haber, Hartel, Shaiu, and
  Wright}]{nci}
Nicholas Sioutos, Sherri~de Coronado, Margaret~W. Haber, Frank~W. Hartel,
  Wen-Ling Shaiu, and Lawrence~W. Wright. 2007.
\newblock \href {https://doi.org/10.1016/j.jbi.2006.02.013} {Nci thesaurus: A
  semantic model integrating cancer-related clinical and molecular
  information}.
\newblock \emph{J. of Biomedical Informatics}, 40(1):30–43.

\bibitem[{Soldaini and Goharian(2016)}]{qumls}
Luca Soldaini and Nazli Goharian. 2016.
\newblock Quickumls: a fast, unsupervised approach for medical concept
  extraction.
\newblock In \emph{MedIR workshop, sigir}, pages 1--4.

\bibitem[{Vaswani et~al.(2017)Vaswani, Shazeer, Parmar, Uszkoreit, Jones,
  Gomez, Kaiser, and Polosukhin}]{transformer}
Ashish Vaswani, Noam Shazeer, Niki Parmar, Jakob Uszkoreit, Llion Jones,
  Aidan~N Gomez, \L~ukasz Kaiser, and Illia Polosukhin. 2017.
\newblock \href
  {http://papers.nips.cc/paper/7181-attention-is-all-you-need.pdf} {Attention
  is all you need}.
\newblock In I.~Guyon, U.~V. Luxburg, S.~Bengio, H.~Wallach, R.~Fergus,
  S.~Vishwanathan, and R.~Garnett, editors, \emph{Advances in Neural
  Information Processing Systems 30}, pages 5998--6008. Curran Associates, Inc.

\bibitem[{Waitzkin(1989)}]{waitzkin1989critical}
Howard Waitzkin. 1989.
\newblock A critical theory of medical discourse: Ideology, social control, and
  the processing of social context in medical encounters.
\newblock \emph{Journal of Health and Social Behavior}, pages 220--239.

\bibitem[{Zhang et~al.(2019)Zhang, Li, Du, Fan, and Philip}]{intent_1}
Chenwei Zhang, Yaliang Li, Nan Du, Wei Fan, and S~Yu Philip. 2019.
\newblock Joint slot filling and intent detection via capsule neural networks.
\newblock In \emph{Proceedings of the 57th Annual Meeting of the Association
  for Computational Linguistics}, pages 5259--5267.

\end{thebibliography}
	\newpage
	\newpage
	\setcounter{figure}{0}
	\renewcommand{\thefigure}{A\arabic{figure}}
	
	\setcounter{table}{0}
	\renewcommand{\thetable}{A\arabic{table}}
	\appendix
	\section*{Appendix}
	\section{Dataset Details}
\label{app:corpus}

\subsection{Dataset Statistics}
\label{app:corpus_stats}
\begin{figure}[h]
    \centering
    \includegraphics[width=\linewidth]{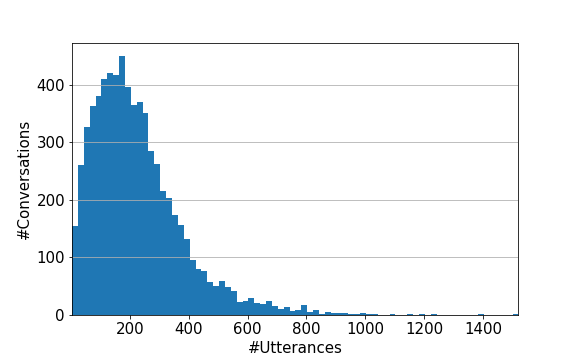}
    \caption{Distribution of the number of utterances in each conversation for the entire dataset.}
    \label{fig:utt_stat}
\end{figure}
The de-identified doctor-patient dialogue corpus used in this work was made available by University of Pittsburgh Medical Center (UPMC) and Abridge AI Inc.. Most of the conversations in this corpus are follow-up encounters between cardiovascular/general medicine doctors and patients. Figure~\ref{fig:utt_stat} shows the distribution of the number of utterances in each conversation. The number ranges from as low as $3$ to as high as $1521$ with a mean of $225$. The proportion of medically relevant utterances in a conversation is quite low (Table~\ref{tab:utt_cat1}). As shown, utterances that belong to the three categories combined make up less than $10\%$ of the conversation portraying the amount of noise present in doctor-patient conversations with regards to further medical processing.
\begin{table}[h!]
	\centering
	\resizebox{0.8\linewidth}{!}{
	\begin{tabular}{lrr}
		\toprule
		\textbf{Category} & \textbf{\#MR-Utt} & \textbf{\#MR-Utt/\#Utt ($\%$)} \\
		\midrule
		Complaints & 6.17 (3.40) & 4.34 (6.06)\\
		Symptoms & 3.56 (4.00) & 1.98 (2.37)\\
		Medications & 4.79 (3.70) & 3.10 (5.49)\\
		\bottomrule
	\end{tabular}
	}
	\captionsetup{font=small,labelfont=small}
	\caption{Avg. (std. dev.) medically relevant utterances (MR-Utt) in each medically relevant category.}
	\label{tab:utt_cat1}
\end{table}

In Table~\ref{tab:position}, we show the average position in the doctor-patient conversation where the speakers start discussing different medical topics. Several of the the encounters are follow-up discussions about a pre-existing complaint. Therefore, patient's current condition with respect to the complaint is often discussed earlier in the conversation. This is generally followed by a discussion about different body systems and associated symptoms that may be bothering the patient, which allows the doctor to prescribe suitable medications. 
\begin{table}[h]
    \centering
    \resizebox{0.6\linewidth}{!}{
    \begin{tabular}{lr}
        \toprule
		\textbf{Category} & \textbf{Relative Position} \\
		\midrule
        Complaints & 0.133 (0.043) \\
        Symptoms & 0.321 (0.057)\\
        Medications & 0.524 (0.069) \\
        \bottomrule
    \end{tabular}
    }
    \captionsetup{font=small,labelfont=small}
    \caption{Avg. (std. dev.) relative position in the conversation where speakers start discussing different medical topics.}
    \label{tab:position}
\end{table}

The above-mentioned flow is merely an ideal depiction of the logical path that could be followed in the dialogue. However, real conversations in the corpus contain multiple topic-switches. For example, discussion of a symptom could be followed by medication which could then lead into a discourse about another symptom and so on.

\begin{table}[h]
\centering
\resizebox{\linewidth}{!}{
\begin{tabular}{L{0.5cm}L{9cm}C{1cm}}
\toprule
 & \textbf{Utterance} & \textbf{Labels} \\ \midrule
 1 & Check if conversation can be added & \\
1 & \textbf{DR: } Good Morning. & \\ \midrule
2 & \textbf{PT: } Good Morning. & \\ \midrule
3 & \textbf{DR: }I'm here with, [PATIENT NAME]. & \\ \midrule
4 & \textbf{DR: }Last time I saw you, you were getting pains in your left leg. Is it still the case? & C,S\\ \midrule
5 & \textbf{PT: }Yes, I do. & S\\ \midrule
6 & \textbf{DR: }Okay, and generally, what are you doing when you get the pains? & \\ \midrule
7 & \textbf{PT: }Um, usually just a heating pad or, you know, ice. & \\ \midrule
8 & \textbf{DR: }Right, but what causes the pains, is what I was getting at? & S\\ \midrule
9 & \textbf{PT: }Uh, I think just the strain of, like, walking, or, or exercise. & S\\ \midrule
10 & \textbf{DR: }All right. & \\ \midrule
11 & \textbf{DR: }I think I am going to ask you to try some Baclofen. & M\\ \midrule
12 & \textbf{DR: }This is a patch you put on the foot when it's bothering you. & M\\ \midrule
13 & \textbf{DR: }Try one patch. & M\\ \midrule
14 & \textbf{DR: }It'll last up to 6 hours. & \\ \midrule
15 & \textbf{PT: }Okay. & \\ \midrule
16 & \textbf{DR: }If you like it, let me know. & \\ \midrule
17 & \textbf{DR: }We'll get you a prescription. & \\ \midrule
18 & \textbf{PT: }Okay. & \\ \midrule
19 & \textbf{DR: }The difference is, you can put this exactly where you need it on the foot and since it's going through the skin, it's not rough on your stomach like, let's say, Ibuprofen or Aspirin or any of the over the counter stuff would be. & \\ \midrule
20 & \textbf{PT: }Okay. & \\
\bottomrule
\end{tabular}
}
\captionsetup{font=small,labelfont=small}
\caption{A constructed example conversation (\textbf{S} = Symptoms, \textbf{C} = Complaints, and \textbf{M} = Medications). Because conversations in the corpus cannot be published or distributed without agreement, the example here is based on a corpus conversation but with the details changed.}
\label{tab:conv}
\end{table}
\subsection{Example Conversation}
\label{app:ex_conv}
An example conversation (details modified) from our corpus is shown in Table~\ref{tab:conv}. Utterances 4,5,8,9 in the conversation discuss a symptom, with the Patient's reply (\ref{tab:conv}:5) acting as an important information about confirmation of the presence of that symptom. Furthermore, \ref{tab:conv}:8 and \ref{tab:conv}:9 provide additional details about the physical activities that cause the symptom. Although, the symptom name is discussed only in \ref{tab:conv}:4, information presented in the other utterances plays an important role in the clinical note. Similarly, utterances \ref{tab:conv}:11,12,13 discuss the medication \textit{Baclofen}. Doctor prescribes the medication in \ref{tab:conv}:11. She provides further information like frequency of usage in \ref{tab:conv}:12, and dosage in \ref{tab:conv}:13, all of which is extremely important for regimen extraction.
\ref{tab:conv}:19 contains names of two medications however it is not a medication utterance. This is case because the utterance does not discuss any medication that the patient is currently taking or being prescribed. The doctor is merely comparing the benefits of her prescribed medication against two popular pain pills.
\subsection{Symptom and Medication Extraction Labels}
In addition to identifying the type of each utterance, corpus annotators also provide a class label to the symptoms and medications from a predefined set. For symptoms, guidelines 
include 178 classes of the form \textit{<Body System>: <symptom>} (e.g. Cardiovascular: Palpitations). Given the small size of the training data, instead of predicting given symptom classes, we predict the body system with which a symptom is associated (Table~\ref{tab:sym_labels}). Table~\ref{tab:sym_targ} contains the list of target UMLS CUIs for each body system that are used as labels for Symptom Extraction. Please note that the list is manually curated and therefore is not exhaustive.
For medications, we manually link each medication label in our training-set to its corresponding UMLS
~\cite{umls} 
concept and group them using hierarchies from
NCI Thesaurus 
\cite{nci} 
(Table~\ref{tab:med_labels}).
\section{Hyperparameters}
\begin{table}[t]
	\centering
	\resizebox{\linewidth}{!}{
	\begin{tabular}{lrr}
		\toprule
		\textbf{Hyper-parameter} & \textbf{Search Range} & \textbf{Best} \\
		\midrule
		GRU hidden size in baselines & $[100, 300, 512]$ & $300$ \\
		\midrule
		Max. utterance length & $[64]$ & $64$ \\
		BERT embedding size & $[768]$ & $768$ \\
		\#Speakers & $[3]$ & $3$ \\
		Speaker embedding size & $[3,4,8,16]$ & $8$ \\
		Number of bins ($k$) & $[4]$ & $4$ \\
		Position embedding size & $[4]$ & $4$ \\
		Semantic Type embedding size & $[8,16]$ & $8$ \\
		BiLSTM hidden size & $[512, 1024, 2048]$ & $1024$ \\
		Weight of $L_{\mathtt{coarse}}$ ($\beta$) & $[0, 0.25, 0.5, 1, 5]$ & $1$ \\
		Learning-rate & $[0.0005, 0.001, 0.01]$ & $0.0005$ \\
		Batch-size & $[8,16,32]$ & $16$ \\
		\bottomrule
	\end{tabular}
	}
	\captionsetup{font=small,labelfont=small}
	\caption{Hyper-parameters. We search over the entire Cartesian product of the different hyper-parameters mentioned here. Best values are chosen using mean AUC of PR curve metric.}
	\label{tab:hp1}
\end{table}
All our experiments are performed on a single Nvidia GeForce GTX 1080 Ti GPU. For \method{} and other BERT-based baselines, we divide the conversations into windows of 128 utterances to ensure fair comparison against BERT-BiLSTM FT, which cannot process more than 128 utterances at a time due to GPU constraints. Other hyperparameters are presented in Table~\ref{tab:hp1}. We perform manual tuning on the entire range of hyper-parameters. AUC under the PR curve metric was chosen to select the best configuration. Results were not very sensitive to different non-zero values of $\beta$.

\section{Utterance Classification}
\subsection{PR Curves}
\begin{figure}[t]
    \centering
    \includegraphics[width=\linewidth]{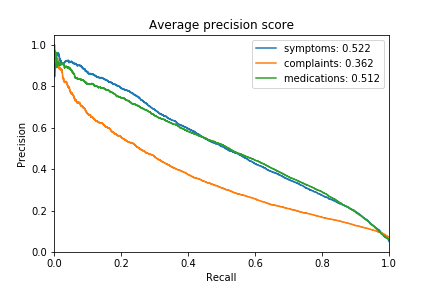}
    \caption{Category-wise PR curves for BERT-FT BiLSTM}
    \label{fig:pr_blstm}
\end{figure}%
\begin{figure}[t]
    \centering
    \includegraphics[width=\linewidth]{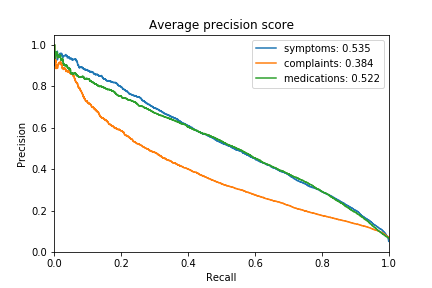}
    \caption{Category-wise PR curves for \method{}}
    \label{fig:pr_multi}
\end{figure}
Figures~\ref{fig:pr_blstm} and \ref{fig:pr_multi} show the precision-recall curves for each category separately. \method{} improves utterance classification for all three categories. For symptom classification, the AUC scores improve from $52.2$ to $53.5$. However, symptom extraction results (Section 6 in the main paper)
suggest that most of this improvement is on identifying utterances that discuss fine-grained details about symptom discussion and not on recognizing the utterance that contains the actual symptom name.

\subsection{Performance on Validation Set}
Table~\ref{tab:val_res} shows the performance of different utterance classification models on the validation set. Similar to the trend shown on test-set, \method{} beats all of the baselines reaching a score of $50.5$ AUC points.
\begin{table}[ht]
	\centering
	\resizebox{\linewidth}{!}{
	\begin{tabular}{lccc}
		\toprule
		\textbf{Methods} & \textbf{Val AUC} & \textbf{\#Param} & \textbf{Time (hrs)} \\
		\midrule
		UniF-BiAGRU & 42.7 & 1.3M & 1\\
		BiF-AGRU & 42.9 & 1.3M & 1\\
		HiGRU-sf & 45.0 & 2.6M & 0.45\\
		\midrule
        BERT & 35.9 & 110M & -\\
		Clinical BioBERT-FT & 38.5 & 110M & 10\\
		BERT-FT & 38.5 & 110M & 10\\
        \midrule
		BERT BiLSTM FT & 47.9 & 125M & 12\\
		BERT-FT BiLSTM & 49.6 & 125M & 10 + 0.1\\
		\midrule
		\method{} & 50.5 & 169M & 10 + 1\\
		\bottomrule
	\end{tabular}
	}
	\caption{Results on val-set and the number of trainable parameters corresponding to each utterance classification model. The time taken by models that use BERT-FT is shown as a sum of two numbers as fine-tuning BERT is only done once, which is then used for both BERT-FT BiLSTM and \method{}.}
	\label{tab:val_res}
	\vspace{-0.3cm}
\end{table}

\section{Downstream Medical Extraction}

\subsection{Micro-F1 vs Threshold}
\begin{figure}[t]
    \centering
    \includegraphics[width=\linewidth]{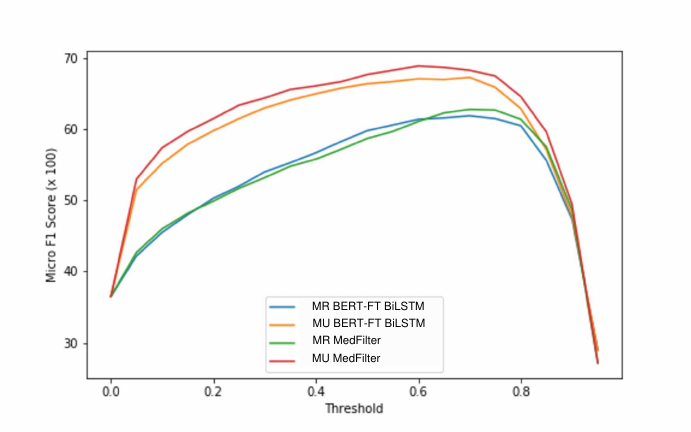}
    \caption{Medication Extraction: Micro-F1 vs Threshold}
    \label{fig:med_ex}
\end{figure}
\begin{figure}[t]
    \centering
    \includegraphics[width=\linewidth]{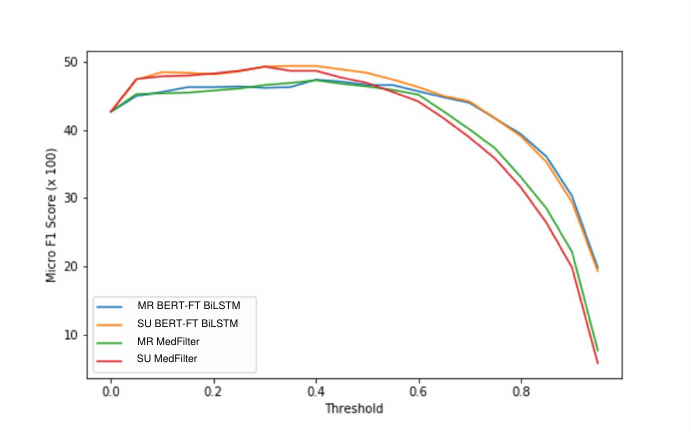}
    \caption{Symptom Extraction: Micro-F1 vs Threshold}
    \label{fig:sym_ex}
\end{figure}
Figure~\ref{fig:med_ex} and ~\ref{fig:sym_ex} show how the performance of medication (ME) and symptom extraction (SE) varies against different utterance topic prediction probability thresholds. We plot the results for BERT-FT BiLSTM and \method{} for brevity.
Micro F1 scores for ME increase monotonically when the threshold is increased from $0$ to $\sim0.75$ (Figure~\ref{fig:med_ex}). This suggests that QuickUMLS medication extractor has low precision that is substantially improved when we prune irrelevant utterances. 
However, the graph for SE (Figure~\ref{fig:sym_ex}) shows that the extractor's performance is dominated by its recall. Pruning helps with improving the precision however does not help with the low recall. This explains the lower gains as compared to ME when using topic-based utterance classification in the SE pipeline (Table 3 in the main paper).
\subsection{Oracle Results}
Table~\ref{tab:medication_or} contains results for medication extraction when medically relevant (MR) or medication (MU) utterances are chosen using an oracle (MR/MU Oracle).
Similarly, oracle results for symptom and complaint extraction are shown in Tables~\ref{tab:symptom_or} and~\ref{tab:comp_or}, respectively. 

We find that there is still a substantial room for improvement in the symptom extraction pipeline. By just improving the topic-based utterance classifier, one can observe a potential jump of 5 Micro-F1 points in symptom extraction. However, we do not observe this trend for medication extraction where the topic-classification done by \method{} performs much better than the Oracle. 

\subsubsection{Why does \method{} perform better than Oracle on Medication Extraction?}
Extraction experiments (like medication extraction or symptom extraction) evaluate the performance at the conversation-level. So, where the medication name gets extracted from within the conversation is irrelevant to the task.

Oracle picks utterances that would be sufficient for a human to identify the medications discussed in the dialogue. However, they might not be adequate for an automatic string-matching based extractor like QuickUMLS. Since QuickUMLS uses non-contextual surface-level features to identify medication names, it would look for phrases (in the input given to it) that match the surface requirements. So, it is possible for the Oracle utterances not to contain the proper surface-level forms that QuickUMLS could leverage for extracting medications. Furthermore, the utterances categorized as medication utterances by MedFilter on the other hand, even though incorrect, might contain the medication names in the form QuickUMLS expects, thus improving the score over the Oracle. One should note, however, that a perfect downstream extractor would not suffer from these side-effects.

\begin{table}[t]
	\centering
	\resizebox{1\linewidth}{!}{
	\begin{tabular}{lcc}
		\hline
		\toprule
		\textbf{Medication Extraction} & \textbf{Macro F1} & \textbf{Micro F1}\\ 
		\midrule
		QuickUMLS (All Text) & 25.4 & 33.5\\
		\midrule
		QuickUMLS (MR Oracle) & 30.0 & 41.3\\
		QuickUMLS (MU Oracle) & 37.6 & 58.3\\
		\midrule
		QuickUMLS (MR \method{}) & 34.2 &62.8\\
		QuickUMLS (MU \method{}) & \textbf{35.9} &\textbf{68.9}\\
		\bottomrule
	\end{tabular}
	}
	\captionsetup{font=small,labelfont=small}
	\caption{Results(\%) for Medication Extraction (ME). \textbf{MR}=Medically Relevant (Symptom + Complaints + Medications) Utterances, \textbf{MU}=Medication Utterances.}
	\label{tab:medication_or}
	\vspace{-0.2cm}
\end{table}
\begin{table}[t]
    \centering
    \resizebox{1\linewidth}{!}{
    \begin{tabular}{lcc}
    \hline
    	\toprule
        \textbf{Symptom Extraction} &\textbf{Macro F1} & \textbf{Micro F1}\\
        \midrule
        QuickUMLS (All Text) & 33.9 & 42.7\\
        \midrule
        QuickUMLS (MR Oracle) & 36.9 & 47.2\\
        QuickUMLS (SU Oracle) & 41.9 & 54.5\\
        \midrule
        QuickUMLS (MR \method{}) & 35.2 & 47.4\\
        QuickUMLS (SU \method{}) & \textbf{36.1} & \textbf{49.3}\\
        \bottomrule
    \end{tabular}
    }
    \captionsetup{font=small,labelfont=small}
    \caption{Results(\%) for Symptom Extraction (SE). \textbf{MR}=Medically Relevant, \textbf{SU}=Symptom Utterances.}
    \label{tab:symptom_or}
\end{table}
\begin{table}[ht]
    \centering
    \resizebox{1\linewidth}{!}{
    \begin{tabular}{lcc}
    \hline
    	\toprule
        \textbf{Complaint Extraction} &\textbf{Macro F1} & \textbf{Micro F1}\\
        \midrule
        QuickUMLS (All Text) & 10.0 & 35.6\\
        \midrule
        QuickUMLS (MR Oracle) & 10.6 & 38.8\\
        QuickUMLS (SU Oracle) & 13.4 & 44.3\\
        \midrule
        QuickUMLS (MR \method{}) & 11.1 & 40.7\\
        QuickUMLS (CU \method{}) & \textbf{11.1} & \textbf{43.7}\\
        \bottomrule
    \end{tabular}
    }
    \captionsetup{font=small,labelfont=small}
    \caption{Results(\%) for Complaint Extraction (CE). \textbf{MR}=Medically Relevant, \textbf{CU}=Complaint Utterances.}
    \label{tab:comp_or}
\end{table}

\subsection{Supervised Extractor}
For symptom extraction (SE), we also show the benefits of using topic-based utterance classification on a supervised-classification based SE approach that leverages a BiLSTM with attention (BiLSTM-Attn) for the problem of predicting the symptoms present in a conversation.
\subsubsection{BiLSTM-Attn}
Each utterance in the conversation is passed through the embedding layer and a BiLSTM layer to obtain a contextualized representation.
\begin{equation*}
\begin{split}
    h_i & = \mathtt{BiLSTM}(e(s_i), h_{i-1}) \\
    \mathbf{H_i} & = \{h_1, h_2, ..., h_n\}
\end{split}
\end{equation*}
where $e(.)$ is the embedding function.
The final state of the BiLSTM is re-weighted using attention calculated as shown in Equation A1.
\begin{equation}
\begin{split}
    h_{final} & = [\cev{h_0}; \vec{h_{n-1}}] \\
    \textbf{S} & = \mathbf{H_i}  h_{final} \\
    \textbf{A} & = softmax(\textbf{S}) \\
    h'_{final} & = \mathbf{H_i^T}\mathbf{A}
\end{split}
\end{equation}
This allows our model to pay attention to important utterances in the conversation to extract symptom information. We pass $h'_{final}$ through a linear classifier and a sigmoid layer to get logits for each possible symptom label (Table~\ref{tab:sym_labels}).

\subsubsection{Experimental Setup}
Similar to the QuickUMLS based extractor, we use Micro and Macro F1 scores to evaluate the performance of the supervised extraction pipeline. BiLSTM-Attn (All Text) model takes in the entire conversation as input, whereas the other variants are given only a subset of utterances. 
MR Oracle/\method{} models are trained on the medically relevant utterances as output by the oracle. Similarly, SU Oracle/\method{} models are trained on the Oracle symptom utterances in each conversation in the training-set. Therefore, topic-based classification is used as a pre-processing step in the pipeline.
\subsubsection{Results}
We present the results for symptom extraction (SE) using a BiLSTM-Attn model in Table~\ref{tab:symptom_or_bi}. We find that using topic-based utterance classification to remove irrelevant utterances before passing the conversation through the BiLSTM-Attn improves the SE performance of the pipeline ($4$ point jump in Micro F1). The results are further improved when the Oracle symptom utterances (SU Oracle) are input to the BiLSTM-Attn.

\begin{table}[h]
    \centering
    \resizebox{1\linewidth}{!}{
    \begin{tabular}{lcc}
    \hline
    	\toprule
        \textbf{Symptom Extraction} &\textbf{Macro F1} & \textbf{Micro F1}\\
        \midrule
        BiLSTM-Attn (All Text) & 28.1 & 57.7 \\
        \midrule
        BiLSTM-Attn (MR Oracle)& 29.3 & 59.4 \\
        BiLSTM-Attn (SU Oracle)& 31.5 & 66.6 \\
        \midrule
        BiLSTM-Attn (MR \method{})& 28.7 & 58.4 \\
        BiLSTM-Attn (SU \method{})& \textbf{29.8} & \textbf{61.7} \\
        \bottomrule
    \end{tabular}
    }
    \captionsetup{font=small,labelfont=small}
    \caption{BiLSTM-Attn results(\%) for Symptom Extraction (SE). \textbf{MR}=Medically Relevant, \textbf{SU}=Symptom Utterances.}
    \label{tab:symptom_or_bi}
\end{table}
\begin{table}[h]
	\centering
	\resizebox{0.35\linewidth}{!}{
	\begin{tabular}{c}
		\toprule
		\multicolumn{1}{c}{\textbf{Symptom Labels}}\\
		\midrule
		Cardiovascular \\
		General \\
		Musculoskeletal \\
		Respiratory \\
		Endocrine \\
		Ear Nose Throat \\
		Eyes \\
		Gastrointestinal \\
		Genital \\
		Head \\
		Neurological \\
		Psychiatric \\
		Skin \\
		Urinary \\
		\bottomrule
	\end{tabular}
	}
	\captionsetup{font=small,labelfont=small}
	\caption{Symptom Extraction Labels (Body Systems)}
	\label{tab:sym_labels}
\end{table}
\begin{table}[h]
	\centering
	\small
	\begin{tabular}{c}
		\toprule
		\multicolumn{1}{c}{\textbf{Chief Complaint Labels}}  \\
		\midrule
		General \\
		Disorder of hematopoietic structure \\
		Disorder of integument, immune system, endocrine \\
		Disorder of musculoskeletal system \\
		Disorder of digestive system \\
		Disorder of the genitourinary system \\
		Disorder of respiratory system \\
		Disorder of breast \\
		Disorder of nervous system \\
		Disorder of cardiovascular system \\
		Others \\
		\bottomrule
	\end{tabular}
	\caption{Complaint labels in the dataset. The label names represent the children of the SNOMED-CT hierarchy: SNOMED CT Concept/Clinical Finding/Finding by site/ Disorder by body site.}
	\label{tab:comp_labels}
\end{table}
\begin{table}[h]
	\centering
	\resizebox{\linewidth}{!}{
	\begin{tabular}{L{3cm}L{7cm}}
		\toprule
		\textbf{Label} & \textbf{Target CUI List}\\
		\midrule
		\textbf{General} & C0036572, C0015672, C0424653, C0015967, C3714552 \\\midrule
        \textbf{Skin} & C0234233, C0178298, C0015230, C0151908 \\
        \textbf{Head} & C0362076, C0042571, C0018681, C0220870, C0012833 \\\midrule
        \textbf{Eyes} & C0235267, C0015397, C0007222, C1705500, C0012634, C2107992, C0017178, C0085635, C0848332, C0521707, C0152227, C0151827, C0017601, C0015230 \\\midrule
        \textbf{Ent} & C0027424, C2926602, C0699744, C0030193, C0031350, C0013456, C0018621, C0009443, C0851354, C0018021, C0017672, C0024117, C0036572, C2012701, C0041912, C0042571, C0019825, C0242429, C0427008, C0497156, C1135208, C0151908 \\\midrule
        \textbf{Genital} & C0567522, C3539891, C3539893, C0149741, C0020624, C2127567, C3539020, C0030193, C0424849, C3539896, C0567523, C0577573, C0007947, C0282005, C0017412, C2129032, C0023533, C4029890, C3539892, C0850758, C0438692, C0567526, C0039591, C0036918, C0036917, C3539890, C0232861, C1657982, C0036916, C3539022, C0877338, C1658964, C1868932, C0423610, C4552766, C0024902, C0234233, C3539023, C0030794, C2075679, C0156398, C1391387, C2030274, C0567519, C0017411, C2032395, C2126231, C0236078, C3539889, C3539895, C0849787, C2032396, C0019693 \\\midrule
        \textbf{Respiratory} & C0857427, C0013404, C0149514, C1396850, C0041312, C0206526, C0019079, C0006277, C0041296, C0010200, C0024115, C0034067, C0030524, C0152874, C0004096, C0041322, C0043144, C0275904 \\\midrule
        \textbf{Cardiovascular} & C0013404, C0795691, C0235710, C0008031, C0035436, C0002871, C0020538, C0018799, C0497234 \\\midrule
        \textbf{Gastrointestinal} & C0011991, C0019196, C0019112, C2032722, C0030193, C0178298, C0854495, C4748517, C0019158, C0018834, C0237938, C0854496, C0849766, C0239549, C0149696, C3553270, C0014724, C0814152, C0000737, C1321898, C0596601, C0085293, C2697368, C0016977, C0949135, C0011226, C0018932, C0017178, C0019159, C0027497, C0687713, C0341286, C0009806, C4728126, C1258215, C0920703, C0019163 \\\midrule
        \textbf{Urinary} & C0392525, C0262655, C0018965, C0239725, C0042029, C0455880, C4087409, C0152032, C0022650, C0021167, C0030193, C0558489 \\\midrule
        \textbf{Musculoskeletal} & C0030193, C0040822, C0858888, C0026857, C1405877, C0158026, C0003864, C0003123, C0030554, C0085593, C0427086, C0426579, C3714552, C0231528, C0036572, C0003873, C0028643, C0424653, C0003862, C0423572, C0427008, C0007859, C0541786, C0522057, C0018099, C2242996, C0015967, C1328469, C0263776, C0015230 \\\midrule
        \textbf{Psychiatric} & C0542476, C1579931, C0235108, C0497307, C0027769 \\\midrule
        \textbf{Neurologic} & C0036572, C0233407, C0042571, C0018681, C0039070, C1660797, C0312422, C0012833, C1135208 \\\midrule
        \textbf{Endocrine} & C0024117, C0020175, C0085602, C0041912, C0848390, C0009443, C0020615, C0221500 \\
		\bottomrule
	\end{tabular}
	}
	\captionsetup{font=small,labelfont=small}
	\caption{CUI Target List for Symptom Extraction}
	\label{tab:sym_targ}
\end{table}

\begin{table}[t]
	\centering
	\resizebox{\linewidth}{!}{
	\begin{tabular}{c}
		\toprule
		\multicolumn{1}{c}{\textbf{Medication Labels}} \\
		\midrule
		DFCBM/Chemical Modifier/Toxin \\
		DFCBM/Dietary Supplement \\
		DFCBM/Drug or Chemical by Structure \\
		DFCBM/Food or Food Product \\
		DFCBM/Industrial Aid \\
		DFCBM/Natural Product \\
		DFCBM/Pharmacologic Substance/Adjuvant \\
		DFCBM/Pharmacologic Substance/AA Blood or Body Fluid \\
		DFCBM/Pharmacologic Substance/AA Cardiovascular System \\
		DFCBM/Pharmacologic Substance/AA Digestive System or Metabolism \\
		DFCBM/Pharmacologic Substance/AA Integumentary System \\
		DFCBM/Pharmacologic Substance/AA Musculoskeletal System \\
		DFCBM/Pharmacologic Substance/AA Nervous System \\
		DFCBM/Pharmacologic Substance/AA Organs of Special Senses \\
		DFCBM/Pharmacologic Substance/AA Respiratory System \\
		DFCBM/Pharmacologic Substance/Anti-Infective Agent \\
		DFCBM/Pharmacologic Substance/Antineoplastic Agent \\
		DFCBM/Pharmacologic Substance/Biological Agent \\
		DFCBM/Pharmacologic Substance/Cation Channel Blocker \\
		DFCBM/Pharmacologic Substance/Chemopreventive Agent \\
		DFCBM/Pharmacologic Substance/Combination Medication \\
		DFCBM/Pharmacologic Substance/Endothelin Receptor Antagonist \\
		DFCBM/Pharmacologic Substance/Enzyme Inhibitor \\
		DFCBM/Pharmacologic Substance/Hormone Therapy Agent \\
		DFCBM/Pharmacologic Substance/Immunotherapeutic Agent \\
		DFCBM/Pharmacologic Substance/Prostaglandin Analogue \\
		DFCBM/Pharmacologic Substance/Protective Agent \\
		DFCBM/Pharmacologic Substance/Protein Synthesis Inhibitor \\
		DFCBM/Physiology-Regulatory Factor \\
		Activity/Clinical or Research Activity/Intervention or Procedure \\
		Manufactured Object/Diagnostic, Therapeutic, or Research Equipment \\
		Others \\
		\bottomrule
	\end{tabular}
	}
	\captionsetup{font=small,labelfont=small}
	\caption{Medication Extraction Labels (DFCBM = Drug, Food, Chemical or Biomedical Material, \textbf{AA} = Agent Affecting).}
	\label{tab:med_labels}
\end{table}

\end{document}